\documentclass[number,preprint]{elsarticle}

\usepackage{booktabs} 
\usepackage{caption} 
\usepackage{subfig} 
\usepackage{subfloat}
\usepackage{amssymb}
\usepackage{amsthm}
\usepackage{amsmath}
\usepackage{color}
\usepackage{epsfig}
\usepackage{times}
\usepackage{url}

\newcommand{\New}{\mathit{New}}
\newcommand{\false}{\mathit{false}}
\newcommand{\true}{\mathit{true}}
\newcommand{\XXa}{{X_{(a)}}}
\newcommand{\XXb}{{X_{(b)}}}
\newcommand{\XXc}{{X_{(c)}}}
\newtheorem{theorem}{Theorem}[section]
\newtheorem{corollary}{Corollary}[section]
\newtheorem{proposition}{Proposition}[section]
\newtheorem{lemma}{Lemma}[section]
\newtheorem{claim}{Claim}[section]
\newtheorem{conj}{Conjecture}[section]
\newtheorem{fact}{Fact}[section]

\newproof{pf}{Proof}
\newdefinition{definition}{Definition}[section]
\newdefinition{example}{Example}[section]

\newcommand{\nop}[1]{}

\newcommand{\ncol}{{N_{\mbox{\rm\tiny\!3COL}}}}
\newcommand{\nkcol}{{N^k_{\mbox{\rm\tiny\!kCOL}}}}

\nop{
\newtheorem{thm}{Theorem}
\newtheorem{conj}[thm]{Conjecture}
\newtheorem{corollary}[thm]{Corollary}
\newtheorem{proposition}[thm]{Proposition}

\newtheorem{lemma}[thm]{Lemma}

\newtheorem{definition}{Definition}

} % END NOP

\journal{Artificial Intelligence}

\begin{document}
\begin{frontmatter}

\title{On Minimal Constraint Networks\tnoteref{t1}}

\tnotetext[t1]{This paper is a significantly extended version of 
a paper with the same title presented at the 17th International Conference on Principles and Practice of 
Constraint Programming~\cite{GGCP11}. The present paper contains new results in addition to those of~\cite{GGCP11}. Possible future updates will be made available on CORR
at~\url{http://arxiv.org/abs/1103.1604}.}

%\author[comlab,omi]{Georg Gottlob}
\author{Georg Gottlob}
\ead{georg.gottlob@cs.ox.ac.uk} 
\address{Department of Computer Science and Oxford Man Institute \\
University of Oxford - Oxford OX1\,3QD,  UK}

\begin{abstract} In a minimal binary constraint network, every tuple of a constraint relation 
can be extended to a solution. The tractability or intractability of computing a solution to 
such a minimal network was a long standing open question. Dechter conjectured this computation problem to be  NP-hard. We prove this conjecture. We also prove a conjecture by Dechter and Pearl stating that for $k\geq 2$  it is NP-hard to decide whether a single constraint  can be decomposed into an
equivalent  $k$-ary constraint network. We show that this holds even in case of bi-valued constraints
where $k\geq 3$, which proves another conjecture of Dechter and Pearl.  Finally, we establish the tractability frontier for this problem with respect to the domain cardinality and the parameter $k$. 
\end{abstract}


\end{frontmatter}

\section{Introduction}

This paper deals with problems related to minimal constraint networks. First, the complexity of computing a solution to a minimal network is determined.  Then, the  problems of
recognizing network minimality and  network-decomposability are studied. 

\subsection{Minimal constraint networks}

In his seminal  1974 paper 
\cite{Mont74}, 
Montanari introduced the concept of a {\it minimal constraint network}. 
Roughly, a minimal network is a constraint network where each partial instantiation
corresponding to a tuple of a constraint relation can be extended to a solution. Each arbitrary binary network $N$ 
having variables $\{X_1,\ldots,X_v\}$ can 
be transformed into an equivalent binary minimal network $M(N)$ by  computing the set $sol(N)$  of all solutions to $N$ and 
creating for $1\leq i<j\leq v$ a constraint $c_{ij}$ whose scope is $(X_i,X_j)$ and whose constraint relation consists 
of  the projection of $sol(N)$ to $(X_i,X_j)$, and for $1\leq i\leq v$ a unary constraint  $c_i$  whose scope is $(X_i)$ and whose constraint relation is the projection of $sol(N)$ over $(X_i)$. The minimal network $M(N)$ is unique, and its solutions are exactly those of the 
original network, i.e., $sol(N)=sol(M(N))$.  

An example of a binary constraint  network $N$ is given in Figure~\ref{MN}a. This network has four variables $X_1,\ldots,X_4$ which, for simplicity, all range over the same numerical domain $\{1,2,3,4,5\}$. Its solution,  $sol(N)$, which is the join of all relations of $N$,  is shown in Figure~\ref{MN}b.  The minimal network $M(N)$ is shown in Figure~\ref{MN}c.

\vspace{0.65cm}

\addtocounter{figure}{-1}
\begin{figure}[h] 
\hrule
\vspace{0.3cm}
\centering  
\subfloat[Binary constraint network {\it N}]{
 {\begin{tabular}{cc} 
  $X_1$&\hspace{-0.1cm}$X_2$   \\ \hline
           $1$&$1$\\
        $1$&$2$ \\ 
        $1$&$3$ \\
     $1$&$5$ \\
  $2$&$1$ \\ 
$2$&$5$ \\ 
$3$&$4$ \\ 
%\phantom{$1$}&\phantom{$1$}\\
  \end{tabular} }
\phantom{aa} \phantom{aa}
{\begin{tabular}{cc} 
  $X_2$    & \hspace{-0.1cm}$X_3$   \\ \hline
           $1$     &  $1$\\
        $1$   & $2$ \\ 
        $2$     &    $2$ \\
$3$&  $1$ \\
  $4$&  $1$ \\ 
 $4$&  $3$ \\ 
 $4$&  $4$ \\ 
%\phantom{$1$}&\phantom{$1$}\\
  \end{tabular}}
\phantom{aa} \phantom{aa}
{\begin{tabular}{cc}  
  $X_1$    & \hspace{-0.1cm}$X_3$   \\ \hline
           $1$     &  $2$\\
        $2$   & $1$ \\ 
        $3$     &    $1$ \\
$3$&  $2$ \\  
$4$&  $1$ \\  
$4$&  $2$ \\  
$4$&  $3$ \\  
%\phantom{$1$}&\phantom{$1$}\\
  \end{tabular} }
\phantom{aa} \phantom{aa}
{\begin{tabular}{cc} 
  $X_3$    & \hspace{-0.1cm}$X_4$   \\ \hline
           $1$     &  $2$\\
        $2$   & $1$ \\ 
$3$   & $1$ \\ 
$3$   & $2$ \\ 
$4$   & $1$ \\ 
$4$   & $2$ \\ 
$4$   & $3$ \\ 
%\phantom{$1$}&\phantom{$1$}\\ 
  \end{tabular}  }
}
\vspace{0.5cm}

\subfloat[Solution $sol(N)$ of {\it N}]{
 {\begin{tabular}{cccc} 
  $X_1$& $X_2$  & $X_3$ & $X_4$ \\ \hline
           $1$&$1$& $2$&$1$ \\
          $1$&$2$& $2$&$1$ \\   
          $2$&$1$& $1$&$2$ \\
          $3$&$4$& $1$&$2$ \\     
%\phantom{.}\\ \vspace{-0.6cm}  
  \end{tabular} }}

\vspace{0.8cm}
\begin{subfigures}
%\subfloat[Minimal network {\it M(N)}]{
\subfloat{}
\subfloat{
 {\begin{tabular}{cc} 
  $X_1$&\hspace{-0.1cm}$X_2$   \\ \hline
        $1$&$1$\\
        $1$&$2$ \\ 
        $2$&$1$ \\
        $3$&$4$ \\  
\phantom{$a_1$}&\phantom{$a_1$}\\
  \end{tabular} }
\phantom{1} 
{\begin{tabular}{cc} 
  $X_2$    & \hspace{-0.1cm}$X_3$   \\ \hline
           $1$     &  $1$\\
        $1$   & $2$ \\ 
        $2$     &    $2$ \\
        $4$&  $1$ \\ 
\phantom{$a_1$}&\phantom{$a_1$}\\
  \end{tabular}}
\phantom{1} 
{\begin{tabular}{cc}  
  $X_1$    & \hspace{-0.1cm}$X_3$   \\ \hline
           $1$     &  $2$\\
        $2$   & $1$ \\ 
        $3$     &    $1$ \\
\phantom{$a_1$}&\phantom{$a_1$}\\
\phantom{$a_1$}&\phantom{$a_1$}\\
  \end{tabular}}
\phantom{1} 
{\begin{tabular}{cc} 
  $X_1$    & \hspace{-0.1cm}$X_4$   \\ \hline
           $1$     &  $1$\\
 $2$     &  $2$\\
$3$     &  $2$\\
\phantom{$1$}&\phantom{$1$}\\
\phantom{$a_1$}&\phantom{$a_1$}\\
  \end{tabular}}
\phantom{1} 
{\begin{tabular}{cc} 
  $X_2$    & \hspace{-0.1cm}$X_4$   \\ \hline
           $1$     &  $1$\\
         $1$     &  $2$\\
  $2$     &  $1$\\
  $4$     &  $2$\\
\phantom{$1$}&\phantom{$1$}\\
  \end{tabular}}
\phantom{1} 
{\begin{tabular}{cc} 
  $X_3$    & \hspace{-0.1cm}$X_4$   \\ \hline
           $1$     &  $2$\\
        $2$   & $1$ \\ 
\phantom{$1$}&\phantom{$1$}\\
\phantom{$a_1$}&\phantom{$1$}\\
\phantom{$1$}&\phantom{$1$}\\
  \end{tabular}}
}
\end{subfigures}

\vspace{-0.5cm}
\subfloat[Minimal network {\it M(N)}]{
%ab hier
%\phantom{a} 
{\begin{tabular}{c} 
  $X_1$    \\ \hline
           $1$ \\
        $2$  \\ 
       $3$ \\
%\phantom{$1$}\\
%\phantom{$1$}\\
\end{tabular}}

\phantom{11} 
{\begin{tabular}{c} 
  $X_2$    \\ \hline
           $1$ \\
        $2$  \\ 
       $4$ \\
%\phantom{$1$}\\
%\phantom{$1$}\\
\end{tabular}}
\phantom{11} 
{\begin{tabular}{c} 
  $X_3$    \\ \hline
        $1$  \\ 
       $2$ \\
%\phantom{$1$}\\
%\phantom{$1$}\\
\phantom{$1$}\\
\end{tabular}}
\phantom{11} 
{\begin{tabular}{c} 
  $X_4$    \\ \hline
        $1$  \\ 
       $2$ \\
%\phantom{$1$}\\
%\phantom{$1$}\\
\phantom{$1$}\\
\end{tabular}}
}

\vspace{0.1cm}

\hrule

\vspace{0.1cm}
\caption{} \label{MN} 
\end{figure}

\pagebreak

Obviously, {\it M(N)}, which can be regarded as an optimally pruned version of 
$N$, is hard to compute. But computing {\it M(N)} may result in a quite useful {\em knowledge compilation}~\cite{Kautz91,Cadoli97}. In fact, with {\it M(N)} at hand, we can answer a number of queries in polynomial time that 
would otherwise be NP-hard.  Typically, these are queries that involve one or two variables only, for example, the queries
"{\em is there a solution for which   $X_4 \leq 3 $?}" or 
"{\em does $N$ have a solution for which $X_2<X_1$?}" are affirmatively answered by a simple lookup in the relevant tables of $M(N)$. For the latter query, for example, one just has to look into the first relation table of $M(N)$, whose tuple $\langle 2,1\rangle$ constitutes a witness.  In contrast, in our example, the query ``{is there a solution for which  $X_1<X_4$?}"  is immediately recognized to have a negative answer, as the fourth relation of {\it M(N)} has no tuple witnessing this inequality.  An example of a slightly more involved non-Boolean two-variable query that can be polynomially answered using {\it M(N)} is: ``{\em what is the maximal value of $X_2$ such that $X_4$ is minimum over all solutions?}". Again, one can just ``read off"  the answer from the single relation of $M(N)$ whose variables are those of the query.  In our example in Figure~\ref{MN}, it is the penultimate relation of {\it M(N)}, that can be easily used to deduce that the answer is $2$ .  

\subsection{Computing solutions to minimal constraint networks}
 
In applications such as computer-supported interactive product configuration, such queries arise frequently, but it would be useful to be able to exhibit at  the same time a 
full solution together with the query answer, that is, an assignment of values to all variables 
witnessing this answer. However, it was even unclear whether the following problem is tractable: 
 Given a nonempty minimal network {\it M(N)}, compute an arbitrary solution to it. Gaur~\cite{Gaur95} formulated this as an open problem. He showed that a stronger version of the problem, 
where solutions restricted by specific value assignments to a pair of variables are sought, is NP-hard, but speculated  that finding 
arbitrary solutions could be tractable. However, since the introduction of minimal networks in 1974, no one came up 
with a polynomial-time algorithm for this task. 
This led Dechter to conjecture that this problem is hard~\cite{Dech03}. Note that this problem deviates in two ways from classical decision problems: 
First, it is a search problem rather than a decision problem, and second, it is a  {\em promise problem}, where it is ``promised" that 
the input networks, which constitute our problem instances, are indeed minimal --- a promise whose verification is itself NP-hard (see Section~\ref{subsec:id1}).  We therefore have to clarify what NP-hardness means, when referring to such problems.  The simplest and probably cleanest definition is the following: The problem is NP-hard if any polynomial algorithms solving it would imply the existence of a polynomial-time algorithm for NP-hard decision problems, and would thus imply P=NP. In the light of this, Dechter's conjecture reads as follows:

\begin{conj}[Dechter\cite{Dech03}]
Unless P=NP,  computing a single solution to a non-empty minimal constraint network cannot be done in polynomial time. 
\end{conj}
 
While the problem has interested a number of researchers, it has not been solved until recently. 
Some progress was made by Bessiere in 2006. In his well-known handbook article ``Constraint Propagation"~\cite{Bess06}, he used 
results of Cros~\cite{Cros03} to show that 
no backtrack-free  algorithm for computing a solution from a minimal network can exist unless the Polynomial Hierarchy collapses to its second level (more precisely, unless $\Sigma^p_2=\Pi^p_2$).  However,  this does not mean that the problem is intractable. 
A backtrack-free algorithm according to Bessiere must be able to recognize {\em each} partial assignment that is extensible to a solution.
In a sense, such an algorithm, even if it computes only one solution, must have the potential to compute all solutions
just by changing the choices of the variable-instantiations made at the different steps. In more colloquial  terms, 
backtrack-free algorithms according to Bessiere must be {\em fair to all solutions}. 
Bessiere's result does not preclude the existence of  a less general 
algorithm  that computes just one solution, while being unable to recognize all partial assignments, and thus being unfair to 
some solutions. 

The simple example in Figure~\ref{MN}, by the way, shows that the following 
na\"{i}ve back\-track-free strategy is doomed to fail: Pick an arbitrary tuple from the first relation of
{\it M(N)}, expand it by a suitable tuple of the second relation, and so on. In fact, if we just picked the first tuple $\langle 1,1\rangle$ of the first relation, we could combine it with the first tuple 
$\langle 1,1\rangle$ of the second relation and obtain the partial instantiation $ X_1=X_2=X_3=1$.
However, this partial instantiation is not part of a solution, as it cannot be expanded to match any tuple of the third relation. While this na\"{i}ve strategy fails, one may still imagine the existence of a more sophisticated backtrack-free strategy, that pre-computes in polynomial time some helpful data structure before embarking on choices.  However, as we show in this paper, such a strategy cannot exist unless NP=P.

In the first part of this paper, we prove Dechter's  conjecture by showing that every polynomial-time search algorithm $A$ that computes a single solution to a minimal network can be transformed into a polynomial-time  decision algorithm $A^*$ for the classical satisfiability problem 3SAT.   The proof is carried-out in Section~\ref{sec:main}.
We first show that each SAT instance can be transformed in polynomial time into an equivalent one that is highly symmetric (Section~\ref{subsec:sym}). Such symmetric instances, which 
we call {\em $k$-supersymmetric}, are then 
polynomially reduced to the problem of computing a solution to a minimal binary constraint network (Section~\ref{subsec:main}). %The minimal networks in the proof, 
%however, have an unbounded number of domain values. 
We further consider the case of bounded domains, that is, when the input instances are such that the cardinality of the 
overall domain of all values that may appear in the constraint relation is bounded by some fixed constant $c$. By a simple modification of the proof of the general case, it is easily seen that even in the bounded domain case, the problem of computing a single solution remains NP-hard (Section~\ref{subsec:bounded}). 

Our hardness results for computing relations can be reformulated in terms of data\-base theory. Every constraint network $N$ can be seen as a relational database instance, where each constraint of $N$ correponds to a single relation instance. The set $sol(N)$ of all solutions to a binary constraint  network (or database instance) $N$ is identical to the relation obtained by performing the natural join of all relation instances of $N$. The minimal network $M(N)$ is then a lossless decomposition of $sol(N)$ according to the join dependency $*[S]$, where $S$ is the schema of $M(N)$. Our main hardness result thus implies that it is coNP hard to recover an arbitrary single tuple of a relation instance $R$ (called a {\em universal relation}) from its lossles decomposition according to a given single join dependency, when only this decomposition is given. Lossles decompositions and universal relations have been studied for many decades, and they were recently related to hidden variable models in quantum mechanics~\cite{Abramsky2,Abramsky1}.  

\subsection{Minimality checking and structure identification}

%In Section~\ref{sec:ident}, we deal with 
%problems of network minimality checking and structure identification. 

In Section~\ref{subsec:id1}, 
we generalize and slightly 
strengthen a result by Gaur~\cite{Gaur95} by showing that 
it is NP-hard to determine whether a $k$-ary network is minimal, 
even in case of bounded domains. 

In Section~\ref{subsec:id2}, 
we study the complexity of  checking whether a 
network $N$ consisting of a single constraint relation 
(typically of arity $\geq k$) can 
be represented by an equivalent $k$-ary constraint network. 
Note that this is precisely the case iff there exists a $k$-ary {\em minimal} network $M$ equivalent to $N$, i.e, one such that $sol(M)=sol(N)$. Dechter and Pearl~\cite{dech-pear-92} 
regarded this problem as a relevant complexity problem of {\em structure identification for relational data}, i.e., of checking whether an element of a 
general class of  objects (in this case, data relations) belongs to a structurally simpler subclass (in this case, $k$-decomposable relations). 
This problem is equivalent to the database problem of testing whether a given instance of a data relation satisfies a specific join dependency.
Dechter and Pearl conjectured that the problem is NP-hard for $k\geq 2$. We prove this conjecture by showing the problem to be coNP-complete for each $k\geq 2$. 

A special case considered in~\cite{dech-pear-92}  is the one of bi-valued constraints, that is, constraints over the Boolean domain. 
For bi-valued constraints, the above structure identification problem is equivalent to testing whether a Boolean formula represented by the explicit list of all its models is equivalent to a $k$-CNF. For $k=2$ this problem is known to be tractable (see~\cite{Dech92,Gaur95}).
Dechter and Pearl~\cite{dech-pear-92} conjectured it to be NP-hard for every fixed $k>2$. In Section~\ref{bi-valued} we prove this conjecture and show that deciding whether bi-valued relations are $k$-decomposable is coNP-complete for each fixed $k>2$. Moreover,  we show in Section~\ref{frontier} that  the representability of tri-valued constraints (and more generally $r$-valued constraints for $r\geq 3$)  as a $k$-ary network is coNP-hard for each fixed $k\geq 2$. Put together, our results allow us to trace the precise tractability frontier for 
the problem of relational structure identification in terms of  the domain cardinality and the parameter $k$.  This is  visualized in Fig~\ref{fig:tracTable} in Section~\ref{frontier}.

\bigskip

The paper is concluded in Section~\ref{sec:conclusion} 
by a brief discussion of the practical significance 
of our main result, a proposal 
for the enhancement of minimal networks, and some hints at possible 
future research.   

\section{Preliminaries and basic definitions}\label{sec:def}
While most  of the definitions in this section are adapted from the standard literature on constraint satisfaction, in particular~\cite{Dech03,Bess06}, we sometimes use a slightly different notation which is more convenient for our purposes.

\noindent
\paragraph{Constraints, networks, and solutions} 
We assume a totally ordered infinite set $({\bf X},\prec)$ of variables. For $X_i,X_j\in {\bf X}$, $X_i\prec X_j$ means that $X_i$ is smaller than $X_j$ according to the "$\prec$" ordering. We assume that all variables of constraint networks are from this set. 
A  {\em k-ary constraint} $c$ is a pair $(scope(c),rel(c))$.  
The scope  $scope(c)$ of $c$  is a sequence $(X_{i_1},X_{i_2}\ldots,X_{i_k})$ of  $k$ distinct variables from $\bf X$, 
where $X_{i_1}\prec X_{i_2}\prec\cdots\prec X_{i_k}$, and where 
each variable $X_{i_j}$  has an associated finite domain  $dom(X_{i_j})$. The relation $rel(c)$ of $c$
is a subset of the Cartesian product  $dom(X_{i_1})\times dom(X_{i_2})\times \cdots\times dom(X_{i_k})$.
%The arity $arity(c)$ of a constraint $c$ is the number of variables in %$scope(c)$. 
The set $\{X_{i_1},\ldots,X_{i_k}\}$ 
of all variables occurring in $scope(c)$ is denoted by $var(c)$. Given that each set of variables is totally ordered by $\prec$, we shall identify each set $U$ of variables, whenever convenient, with the list $\vec{U}$ of its elements ordered according to $\prec$. We thus may write $scope(c)=U$ instead of $scope(c)=\vec{U}$. More generally, since the concepts of lists of distinct elements  and ordered sets coincide, we may use set-theoretic notatation to express facts about    
such lists. For example, if $s$ denotes a scope, we may write $X_i\in s$ to express that $X_i$ is a variable in this scope. If, moreover,  $U$ denotes  a  list (or even an unordered set) of variables, we may write $s\subseteq U$ to say that each variable of $s$ is also an element of $U$, and so on.

A {\it Constraint Network} $N$ consists of  a finite set $var(N)=\{X_1,\ldots,X_v\}$ of variables with associated domains $dom(X_i)$ for $1\leq i\leq v$, and a set of constraints $cons(N)=\{c_1,\ldots,c_m\}$, where for $1\leq i\leq m$,  $var(c_i)\subseteq var(N)$.

If $U\subseteq var(N)$ is a set of variables, then $dom(U)=\bigcup_{X\in U}dom(X)$. The {\it domain} $dom(N)$ of a constraint network $N$ is defined by $dom(N)=dom(var(N))$. The {\it schema} of $N$ is the set 
$schema(N)=\{scope(c)|c\in cons(N)\}$
of all scopes of the constraints of $N$. If $S$ is a schema, 
then $var(S)$ denotes the set of all variables in the scopes of $S$. In particular, if $S$ is the schema of network  $N$, we have $var(S)=var(N)$. We call $N$ {\em binary} ($k$-{\it ary}) if  $arity(c)\leq 2$\,  ($arity(c)\leq k$) for each constraint $c\in cons(N)$.

Let $N$ be a constraint network. An {\it instantiation mapping} for a set of variables $W\subseteq var(N)$ is a mapping 
$\theta:W\longrightarrow dom(W)$, such that for each $X\in var(N)$, $\theta(X)\in dom(X)$. 
We call $\theta(W)$ an instantiation of $W$. An instantiation of a proper subset $W$ of $var(N)$ is called a {\em partial instantiation} while 
an instantiation of $var(N)$ is called a {\em full instantiation} (also {\em total instantiation}). A constraint $c$ of $N$ is {\em satisfied} 
by an instantiation mapping $\theta:W\longrightarrow dom(W)$ if whenever $var(c)\subseteq W$, then 
$\theta(scope(c))\in rel(c)$.   An instantiation mapping  $\theta:W\longrightarrow dom(W)$  is {\em consistent} if it is satisfied by all constraints.  
By abuse of terminology, if  $\theta$ is understood and is consistent, then we may also say that 
$\theta(W)$ is consistent. A {\em solution} to a constraint network $N$ is a consistent full instantiation for $N$. 
The set of all solutions 
of $N$ is denoted by $sol(N)$. $N$ is {\em solvable} iff $sol(N)\neq \emptyset$. Whenever useful, we will identify the solution set $sol(N)$ with a single constraint whose scope is $var(N)$
and whose relation consists of  all tuples in $sol(N)$. 
We assume without loss of generality, that for each set of variables $W\subseteq var(N)$ of a constraint network, there 
exists at most one constraint $c$ such that $var(c)=W$. (In fact, if there are two or 
more constraints with exactly the same variables in the scope, an equivalent single  constraint can always be obtained by intersecting 
the constraint relations.)

\smallskip

\noindent
\paragraph{Complete networks} 
The {\em complete schema} $S^U_k$ over a set of variables $U$ denotes the schema consisting of all nonempty constraint scopes of arity at most $k$ contained in $U$. For example, if $U=\{X_1,X_2\}$, then 
$S^U_k=\{(X_1), (X_2), (X_1,X_2)\}$. If the set of variables $U$ is understood, we will write $S_k$ instead of $S^U_k$.
A $k$-ary constraint network $N$ is {\em complete}, if its schema is $S^{var(N)}_k$. For each fixed constant $k$, each $k$-ary constraint network $N$ can be transformed 
by a trivial polynomial reduction into an equivalent complete $k$-ary network $N^+$ with 
$sol(N)=sol(N^+)$. In fact, if $\ell\leq k$, then for each (ordered) set of variables  $W=\{X_{i_1},\ldots,X_{i_\ell}\}$ that is in no scope of $N$, we may just add the trivial 
constraint $\top_W$ with $scope(\top_W)=(X_{i_1},\ldots,X_{i_\ell})$ and $rel(\top_W)=dom(X_{i_1})\times dom(X_{i_2})\times\cdots\times dom(X_{i_\ell})$.  
For this reason, we may, whenever useful, restrict our attention to complete networks.
Some authors, such as Montanari~\cite{Mont74} who studies binary networks, assume by definition that all networks are complete, others, such as Dechter~\cite{Dech03} make this assumption implicitly. 
\nop{We here assume unless otherwise stated, 
that $k$-ary networks are complete. 
%but we will often not write out the trivial constraints $\top_{ij}$, but will simply assume they exist in case no other constraint between $X_i$ and 
%$X_j$ is specified. 
In particular, we will assume without loss of generality, that when a binary constraint network $N$ is defined over 
variables $X_1,\ldots,X_v$, that are given in this order, then the constraints are such that their scopes are precisely 
all pairs $(X_i,X_j)$ such that $1\leq i<j\leq v$.  For a binary constraint network $N$ over variables $\{X_1,\ldots,X_v\}$, we denote the constraint
with scope  $(X_i,X_j)$ by $c^N_{ij}$.}

\smallskip
\paragraph{Intersections of networks, containment, and projections}
Let $N_1$ and $N_2$ be two constraint networks defined over the same schema $S$ (that is, the same set $S$ of constraint scopes). 
The {\em intersection} $M=N_1\cap N_2$ of $N_1$ and $N_2$  is the network having $var(M)=var(N_1)=var(N_2)$, and having a constraint $c^s$, for each $s\in S$, such that $scope(c^s)=s$ and  
$rel(c^s)=rel(c_1^s)\cap rel(c_2^s)$, where $c_1$ and $c_2$ are the constraints having scope $s$ of 
$N_1$ and $N_2$, respectively.   The intersection of arbitrary families of constraint networks defined over 
the same schema is defined in a similar way. For two networks $N_1$ and $N_2$ over the same 
schema $S$, we say that $c_1$ is {\it contained in} $c_2$, and write $N_1\subseteq N_2$, 
if for each $s\in S$, and for $c_1\in cons(N_1)$ and $c_2\in cons(N_2)$ 
with $scope(c_1)=scope(c_2)=s$, $rel(c_1)\subseteq rel(c_2)$. If $c$ is a constraint over a set of variables $W=\{X_1,\ldots,X_v\}$ and $V\subseteq W$, then the projection $\Pi_V(c)$ is the constraint 
whose scope is $V$, and whose relation is the projection over $V$ of $rel(c)$.
Let $c$ be  a constraint and  $S$ a schema consisting of one or more scopes contained 
in $scope(c)$, then $\Pi_S(c)=\{\Pi_s(c)|s\in S\}$. 
\nop{If $N$ is a constraint network and 
$S$ a schema all of whose variables are contained in $var(N)$, then $\Pi_S(N)$ is the constraint network over schema $S$  whose set of constraints is $\{\Pi_S(c)|c\in N\}$.  }

\smallskip

\paragraph{Minimal networks}
 Let $c$ be a constraint with $var(c)=U$. The projection  $\Pi_{S^U_k}(c)$ will henceforth just denote by 
$\Pi_{S_k}(c)$. Thus $\Pi_{S_k}(c)$ is the constraint network 
obtained by projecting $c$ over all scopes in the schema $S^U_k$ (simply denoted by $S_k$), i.e., over all  nonempty ordered lists of at most $k$ variables from $var(c)$.  In particular, the constraints of $\Pi_{S_2}(c)$ are precisely all $\Pi_W(c)$ such that $W\subseteq var(c)$ is a unary or binary scope.

It was first observed in~\cite{Mont74} that for each binary constraint network $N$, there is a unique binary {\em minimal network} $M(N)$ that consists of the intersection of all binary networks $N'$  over schema $S_2$ for which $sol(N')=sol(N)$. Minimality here is with respect to the above defined ``$\subseteq$''-relation among binary networks. More 
generally, for each $k$-ary network $N$ there is a unique $k$-ary minimal network 
$M_k(N)$ that is the intersection of all $k$-ary networks $N'$  over schema $S_k$ for which $sol(N')=sol(N)$.
(For the special case $k=2$ we have $M_2(N)=M(N)$.)
The following is well-known \cite{Mont74,MontRoss88,Dech03,Bess06,Honeyman} and easy to see: 
\begin{itemize}
\item $M_k(N) =\Pi_{S_k}(sol(N))$.
\item $M_k(N)\subseteq N'$ for all $k$-ary networks $N'$ with $sol(N')=sol(N).$
\item A $k$-ary network $N$ is satisfiable (i.e., has at least one solution) iff $M_k(N)$ is non-empty.
\item A $k$-ary network $N$ is minimal iff $\Pi_{S_k}(sol(N))=N$.
\item A $k$-ary network $N$ is minimal iff $M_k(N)=N$.
\item A network $N$ over schema $S_k$ is minimal  iff  there exists a {\em universal relation} $\rho$ for $N$, that is, a single constraint $\rho$  such that $N=\Pi_{S_k}(\rho)$.  In this case $N$ is said to be {\em join consistent} (see~\cite{Honeyman}).
\end{itemize}

It is obvious that for $k\geq 2$, $M_k(N)$  is hard to compute. In fact, just {\em deciding} whether for a network $N$,  $M_k(N)$ is the empty network is coNP-complete, because 
this decision problem is equivalent to deciding whether $N$ has no solution. (Recall that deciding whether a network $N$ has a solution is 
NP-complete~\cite{MackFreu85}.)\,  In this paper, however, we are not primarily interested in computing $M_k(N)$, but in computing a single solution, in case  
$M_k(N)$ has already been computed and is known. 

\paragraph{Graph theoretic characterization of minimal networks}\,	
An {\it $n$-partite graph} is a graph whose vertices can be partitioned into $n$ disjoint sets so that 
no two vertices from the same set are adjacent. It is well-known (see, e.g.,~\cite{Tsang93}) that  
each binary constraint network $N$ on $n$ variables can be represented as an $n$-partite graph $G_N$. The 
vertices of $G_N$ are possible instantiations of the variables by their corresponding domain values. Thus, 
for each variable $X_i$ and possible domain value $a\in dom(X_i)$, there is a vertex $X_i^a$. Two vertices $X_i^a$ and $X_j^b$ are 
connected by an edge in $G_N$ iff the relation of the constraint $c^N_{ij}$ with scope $(X_i,Y_j)$ contains the tuple $(a,b)$.\footnote{We disregard unary relations of $N$ here; in fact, each unary relation of a constraint network can be eliminated by appropriately restricting the domain of its scope variable.} Gaur~\cite{Gaur95} gave the following nice characterization of minimal networks: A solvable\footnote{We here refer to solvability according to our definition; Gaur uses a different definition of this term.} complete binary constraint network $N$ on $n$ variables is minimal iff each
edge of $N$ is part of a clique of size $n$ of $G_N$. Note that by definition of $G_N$ as an $n$-partite graph, there cannot be any 
clique in $G_N$ with more than $n$ vertices, and thus the cliques of $n$ vertices are precisely the 
maximum cliques of $G_N$.   

\smallskip 

\noindent
\paragraph{Satisfiability problems}\  An instance $C$ of the {\em Satisfiability (SAT)} problem is a conjunction of clauses (often 
just written as a {\em set} of clauses), each of which consists of a disjunction ({\em often written as set}) of literals, i.e., of positive or negated  {\em propositional variables}. Propositional variables are also called {\em (propositional) atoms}. 
If $\alpha$ is a set of clauses or a single clause, then we denote by $propvar(\alpha)$ the set of all propositional 
variables occurring in $\alpha$. 

A 3SAT instance is a SAT instance each clause of which is a disjunction of at most three literals. 3SAT is the problem of deciding whether a 3SAT instance is satisfiable.

\section{NP-hardness of computing minimal network solutions}
\label{sec:main}

To show that computing a single solution to a minimal network is NP-hard, we will do exactly the contrary of what people
--- or automatic constraint solvers ---  usually do whilst solving a constraint network or a SAT instance. 
While everybody aims at breaking symmetries, we will actually {\em introduce additional symmetry} into a 3SAT instance and its corresponding constraint network
representation. This will be achieved by the {\it Symmetry Lemma} to be proved in the next section.

\subsection{The Symmetry Lemma}\label{subsec:sym}

The following lemma shows  that, for each fixed $k\geq 1$,  
one can  transform an arbitrary 3SAT instance $C$ in polynomial time into a satisfiability-equivalent 
highly symmetric  SAT instance $C^*$ such that, whenever $C$ (and thus $C^*$) is satisfiable, each truth value assignment to any $k$ variables of $C^*$ can be extended to a truth value assignment satisfying $C^*$.  Before stating the lemma, let us formally define this notion of symmetry, which we refer to as  {\em supersymmetry}.

\begin{definition}
For $k\geq 1$, a SAT instance $C$ is $k${\em -supersymmetric} if $C$ is either unsatisfiable 
or if for each set of $k$ propositional variables $\{p_1,\ldots, p_k\}\subseteq propvar(C)$ and for each arbitrary truth value assignment $\eta$  to $\{p_1,\ldots,p_k\}$, 
there exists a satisfying truth value assignment 
$\tau$ for $C$ that extends $\eta$. A SAT instance that is  $2$-supersymmetric is also called {\em supersymmetric}.
\end{definition}

\smallskip 

Assume $k<k'$. By the above  definition,  if  a SAT instance $C$ is $k'$-supersymmetric, then $C$ is also $k$-supersymmetric. However, 
a $k$-supersymmetric SAT  instance $C$ is not necessarily also $k'$-supersymmetric. 

\smallskip

\begin{lemma}[Symmetry Lemma]~\label{lem:sym}
For each fixed integer $k\geq 1$, there is a polynomial-time transformation  $T$
that transforms each 3SAT instance $C$  
into a $k$-supersymmetric SAT instance $C^*$ such that $C$ is satisfiable iff $C^*$ is satisfiable. 
\end{lemma}

We  illustrate the proof of Lemma~\ref{lem:sym} by an example. A full proof is given in~\ref{AppendixA}

\begin{proof} {\em (Illustration by Example)}
Consider the 3SAT instance   $C=C_1\wedge C_2\wedge C_3$, where 
 
$$
\begin{array}{ccccccc}
C_1& = & \phantom{\neg}p & \vee &  \neg q  & \vee &  r\\
C_2& = & \neg p & \vee &  \neg q  & \\
C_3& = &          &  &  \phantom{\neg} q  &   &   \\
\end{array} 
$$

Clearly, the above 3SAT instance $C$, while satisfiable,  is not even $1$-supersymmetric, and therefore, a fortiori, not $k$-supersymmetric for any $k\geq 1$. To see this, observe that the  partial truth-value assignment assigning  $\false$ to $q$ always falsifies clause $C_3$, and can thus not be extended to a satisfying truth value assignment for $C$. In the sequel, we illustrate how $C$ can be transformed 
by a polynomial-time transformation $T$ into a satisfiable supersymmetric SAT instance $C^*=T(C)$. To this aim we introduce to each propositional variable $v$ of $C$ a set $\New(v)$ of five new propositional variables. In particular, we have: 

 $$
\begin{array}{ccll}
\New(p)& = &  \{p_1,p_2,p_3,p_4,p_5\},\\
\New(q)& = &  \{q_1,q_2,q_3,q_4,q_5\}, \ \mbox{\normalfont and}\\
\New(r)& = &  \{r_1,r_2,r_3,r_4,r_5\}.
\end{array} 
$$

We now create $C^*$ from $C$ by taking the conjunction of all clauses obtained by replacing in each clause of $C$ each positive literal  $v$ in all possible ways by the disjunction $v_i \vee v_j\vee v_k$ of three elements $v_i, v_j,v_k\in \New(v)$, and by replacing each negative literal $\neg v$ in all possible ways by the disjunction $\neg v_i \vee \neg v_j\vee \neg v_k$, 
where $v_i, v_j,  v_k$ are elements of $\New(v)$. Each clause is thus replaced by a multitude of other clauses that are all taken in conjunction. In particular, in our example, clause $C_1$ will actually be replaced by the conjunction of the following 1000 clauses $C_1^1\ldots C_1^{1000}$: 

$$
\begin{array}{ccccccc}
C_1^1:& p_1 \vee  p_2   \vee p_3 & \vee & \neg q_1  \vee  \neg q_2  \vee \neg q_3 & \vee &
r_1  \vee r_2  \vee  r_3 ;\vspace {0.2cm}\\

C_1^2:& p_1 \vee  p_2   \vee p_3 & \vee & \neg q_1  \vee  \neg q_2  \vee \neg q_3 & \vee &
r_1  \vee r_2  \vee  r_4;\vspace {0.2cm}\\

C_1^3:& p_1 \vee  p_2   \vee p_3 & \vee & \neg q_1  \vee  \neg q_2  \vee \neg q_3 & \vee &
r_1  \vee r_2  \vee  r_5;\vspace {0.05cm}\\

\ldots & \ldots\ldots & . & \ldots\ldots &   . & \ldots\ldots \vspace {0.05cm}\\

C_1^{10}:& p_1 \vee  p_2   \vee p_3 & \vee & \neg q_1  \vee  \neg q_2  \vee \neg q_3 & \vee &
r_3  \vee r_4  \vee  r_5;\vspace {0.2cm}\\

C_1^{11}:& p_1 \vee  p_2   \vee p_3 & \vee & \neg q_1  \vee  \neg q_2  \vee \neg q_4 & \vee &
r_1  \vee r_2  \vee  r_3;\vspace {0.05cm}\\

\ldots & \ldots\ldots & . & \ldots\ldots &   . & \ldots\ldots \vspace {0.05cm}\\

C_1^{1000}:& p_ 3\vee  p_4   \vee p_5 & \vee & \neg q_3  \vee  \neg q_4  \vee \neg q_5 & \vee &
r_3  \vee r_4  \vee  r_5.\\
\end{array} 
$$

\medskip

\noindent 
Similarly, clause $C_2=  \neg p \vee   \neg q$ is replaced by the following 100 clauses $C_2^1\ldots  C_2^{100}$:

$$ 
\begin{array}{ccccc}
C_2^1:& \neg p_1 \vee \neg  p_2   \vee\neg  p_3 & \vee &  \neg q_1  \vee \neg q_2  \vee  \neg q_3: & \vspace {0.2cm}\\
C_2^1:& \neg p_1 \vee \neg  p_2   \vee\neg  p_3 & \vee &  \neg q _1  \vee \neg q_2  \vee  \neg q_4;& \vspace {0.05cm}\\
\ldots & \ldots\ldots & . & \ldots\ldots &   \vspace {0.05cm}\\
C_2^{100}:& \neg p_3 \vee \neg  p_4   \vee\neg  p_5 & \vee &  \neg q_3  \vee \neg q_4  \vee  \neg q_5. &\\
\end{array} 
$$
\medskip

\noindent
Finally, clause $C_3=\,p$ is replaced by the following 10 clauses $C_3^1,\ldots,C_3^{10}$\ :

$$ 
\begin{array}{cc}
C_3^1:&  q_1 \vee  q_2   \vee  q_3;  \vspace {0.2cm}\\
C_3^2:&  q_1 \vee  q_2   \vee  q_4;  \vspace {0.05cm}\\
\ldots & \ldots\ldots  \vspace{0.05cm}\\
C_3^{10}:&  q_3 \vee  q_4   \vee q_5. \\
\end{array} 
$$

\medskip

\noindent
The SAT instance $C^*=T(C)$ then consists of the conjunction of all these clauses: 
$$C^*\,=\,C_1^1\wedge \ldots\wedge C_1^{1000}\wedge C_2^1\wedge\ldots\wedge  C_2^{100}\wedge C_3^1\wedge\ldots\wedge C_3^{10}.$$ 

\medskip

We claim  ---\,and formally prove in~\ref{AppendixA}\,--- that the above transformation from a 3SAT instance $C$ to a SAT instance $C^*$ satisfies the following two key facts: 
\begin{description}
\item[Fact 1] {\em $C^*$ is satisfiable iff $C$ is satisfiable\ (in our example, $C^*$  is thus satisfiable).}  In fact, each satisfiable truth value assignment $\tau$ to the propositional variables of $C$ can be transformed to a satisfying truth value assignment $\tau^*$  to $C^*$ as follows: 
If $\tau(v)=\true$, then let $\tau^*$ assign $\true$ to at least three propositional variables in $\New(v)$, and $\false$ to all others, and if 
$\tau(v)=\false$, then let $\tau^*$ assign $\false$ to at least three  propositional variables in $\New(v)$, and $\true$ to all others. In our example, for instance, consider the truth value assignment $\tau$ satisfying $C$, where $\tau(p)=\false$ and 
$\tau(q)=\tau(r)=\true$. This truth value assignment satisfies $r$ and therefore $C_1$. The assignment $\tau^*$ to $C^*$ thus assigns $\true$ to at least three atoms from $\New(r)=\{r_1,r_2,r_3,r_4,r_5\}$, assume, for example to  $\{r_1,r_4,r_5\}$. But each  3-element subset of $\New(r)$  has a non-empty intersection with each other non-empty three element subset of $\New(r)$, and thus with the set of atoms of each and every clause $C_1^i\in \{ C_1^1,\ldots,C_1^{1000}\}$. Therefore,  each such clause   $C_1^i$ is satisfied. For example,  $C_1^1$ has a $r_1$ in common with the set   $\{r_1,r_4,r_5\}$, and so must be satisfied 
by $\tau^*$ for $\tau^*(r_1)=\true$. A similar argument holds for negative literals. Applying the same type of reasoning to all clauses  $C_i$ of $C$, 
given that each such $C_i$ has at least one literal satisfied by  $\tau$, all clauses $C_i^j$  of $C^*$ are satisfied by $\tau^*$. In summary, $\tau^*$ satisfies $C^*$. Vice versa, we  show in the full proof  that if $C^*$ is satisfiable, then so must be  $C$.

\item[Fact 2] $C^*$ {\em is supersymmetric}. Intuitively, this is due to the great choice  of truth value assignments to the propositional variables in $\New(v)$, when constructing 
a satisfying assignment $\tau^*$ for $C^*$, as above, from an assignment $\tau$ for $C$. Imagine, for illustration, we'd like to construct a truth value assignment $\tau^*$  satisfying our example-instance  $C^*$, such that $\tau^*(p_1)=\true$ and $\tau^*(q_3)=\false$. Note that no truth value assignment to instance $C$ can actually satisfy $p$ or falsify $q$. Notwithstanding, we are  able to find an appropriate $\tau^*$ with the desired properties. We start with an arbitrary satisfying truth value assignment $\tau$ to $C$, for example, the one where $\tau(p)=\false$ and 
$\tau(q)=\tau(r)=\true$. To construct $\tau^*$, let us first define  $\tau^*$ on  the elements of $\New(p)=\{p_1,\ldots,p_5\}$. 
According to the construction rules for $\tau^*$  in the previous paragraph, given that $\tau(p)=\false$,  $\tau^*$ must  assign  $\false$ to at least three elements of $\New(p)$, but not necessarily to all elements of $\New(p)$.  This leaves us the freedom of assigning $true$ to $p_1$. So, we can, for example,  assign $\false$ to $p_2,p_3,p_4,$ and $p_5$, and $\true$ to $p_1$. Similarly, given that $\tau(q)=true$, $\tau^*$ must assign $\true$ to at least three elements of $\New(q)$, which can be done while fulfilling at the same time our requirement  that $\tau^*(q_3)=\false$. For example, let $\tau^*(q_1)=\tau^*(q_2)=\tau^*(q_5)=\true$ and  $\tau^*(q_3)=\tau^*(q_4)=\false$. 
Finally, the only requirement regarding the truth values assigned by $\tau^*$ to  the elements of $\New(r)$  is that at least three of these propositional variables be assigned $\true$.  Thus,  for example, let $\tau^*(r_1)=\tau^*(r_2)=\ldots=\tau^*(r_5)=\true$. In summary, it is easy to see (and actually follows from Fact 1) that the truth value assignment $\tau^*$ constructed this way satisfies $C^*$. Moreover, $\tau^*$ extends the initially given partial truth value assignment $\tau^*(p_1)=\true$ and $\tau^*(q_3)=\false$. More generally,  for every  pair $v,w$ of propositional variables of $C^*$, and for every truth value assignment $\eta$ to $\{v,w\}$,  one can construct a truth value assignment $\tau^*$ that extends $\eta$ and satisfies $C^*$
This shows that $C^*$ is 2-supersymmetric, i.e., supersymmetric.
\end{description}

As easily seen,  the transformation from an arbitrary 3SAT instance $C$ to the corresponding $C^*$  is polynomial-time computable. Together with Facts~1 and~2, this informally proves the Symmetry Lemma for $k=2$. For  $k>2$, the proof is analogous. 
\end{proof}
%\smallskip

\noindent 
REMARK. The concept of supersymmetry is somewhat related to the notions of {\em quadrangle} and {\em subquadrangle} defined in~\cite{Rodo96} and further discussed in~\cite{Cohe-Houg-05}. A quadrangle is a single constraint $c$ that is satisfied for all  value-assignments that assign any arbitrary value from $dom(X)$ to each variable $X$ in $scope(c)$. Thus, the constraint relation $rel(c)$ of a quadrangle $c$ simply consists of a Cartesian product of domains. An $n$-ary constraint $c$ is a subquadrangle if each projection of $c$ to $n-1$ or fewer variables from $scope(c)$ is a quadrangle. Generalizing this notion, we define a $k$-{\em subquadrangle} to be a constraint, all of  whose projections to $k$ variables are quadrangles. In this context, Lemma~\ref{lem:sym} may be reformulated as follows: For each $k\geq 1$, every satisfiable 3SAT instance $C$ can be transformed to a satisfiable SAT instance $C^*$ whose solution relation $sol(C^*)$ is a $k$-subquadrangle.

\medskip

\subsection{Intractability of computing solutions}\label{subsec:main}

\medskip

The Symmetry Lemma is used for proving our main intractability result.

\begin{theorem} \label{theo:main}
For each fixed constant $k\geq 2$, unless NP=P,  computing a 
single solution from a minimal $k$-ary constraint network $N$ cannot be done in polynomial time. The 
problem remains intractable even if the cardinality of each variable-domain is bounded by 
a fixed constant. 
\end{theorem}

\noindent{\bf Proof.} \   We first prove the theorem for $k=2$. 
Assume $A$ is an algorithm that computes in time $p(n)$, where $p$ is some polynomial,  
a solution $A(N)$ to each non-empty 
minimal binary constraint network $N$ of size $n$\nop{, for which there exists some $N^+$ as described}. 
We will construct a polynomial-time 3SAT-solver $A^*$ from $A$. The theorem then follows. 

Let us first define a simple transformation $S$ from SAT instances to equivalent  binary constraint networks. 
$S$ transforms  conjunctions $K\, = \,  K_1\, \wedge \cdots\wedge\,K_r$ of at least two clauses into binary constraint  networks $S(K)=N_K$ as follows. 
The set of variables $var(N_K)$ is defined by $var(N_K)=\{K_1,\ldots,K_r\}$. For each variable $K_i$ of $N_K$, 
the domain $dom(K_i)$ consists exactly of all literals appearing in $K_i$. For each distinct pair of clauses $(K_i,K_j)$, 
$i<j$, there is a constraint $c_{ij}$ having $scope(c_{ij})=(K_i,K_j)$ and $rel(c_{ij})=(dom(K_i)\times dom(K_j))-\{(p,\bar{p}), (\bar p,p)\, |\, p\in propvar(K)\}$. Moreover, for each variable $K_i$ there is a unary constraint $c_i$ whise scope contains the single variable $K_i$,  such that $rel(c_i)=dom(K_i)$.
It is easy to see that $K$ is satisfiable iff $N_K$ is solvable. Basically, $N_K$ is solvable, iff we can pick one literal per clause such that 
the set of all picked literals contains no atom together with its negation. But this is just equivalent to the satisfiability of $K$. The transformation $S$ is
clearly polynomial-time computable.

Now consider constraint networks $N_{C^*}=S(C^*)$, where $C^*$ is obtained via transformation $T$ as in Lemma~\ref{lem:sym} from 
some 3SAT instance $C$, i.e., $C^*=T(C)$.   In a precise sense, $N_{C^*}$ inherits the high symmetry present in $C^*$. In fact, if $C^*$ is 
satisfiable, then, by Lemma~\ref{lem:sym}, for every pair $\ell_1,\ell_2$ of non-contradictory literals, there is a satisfying assignment that 
makes both literals true. Thus, if $C^*$ (and thus $C$)  is satisfiable,  for every constraint $c_{ij}$, 
we may pick each 
pair $(\ell_1,\ell_2)$ in $rel(c_{ij})$ as part of a solution, and thus no such pair is useless. Moreover, if $C^*$ is satisfiable, then, clearly, each value in the relations of each unary constraint $c_i$ is part of a solution.
It follows that if $C^*$ --- and thus $C$ --- is satisfiable, then $M(N_{C^*})=N_{C^*}$, which means that $N_{C^*}$ is minimal. We thus have:
\begin{quote}
\begin{description}
\item[$(*)$] If $C$ is satisfiable then $N_{C^*}$ is non-empty and minimal.
\end{description}
\end{quote}
%On the other hand, if $C^*$ (and thus $C$) is not satisfiable, then 
%$M(N_{C^*})$ is the empty network. 
%Thus $C$ is satisfiable iff $M(N_{C^*})=N_{C^*}$ i.e., iff $N_{C^*}$ is minimal~\footnote{From this, 
%by the way, it follows that checking whether a 
%given binary network is minimal is NP-hard, and thus NP-complete; see also Section~\ref{subsec:id1}.}  

\begin{figure*}[h]
\centering
\includegraphics[width=0.8\textwidth]{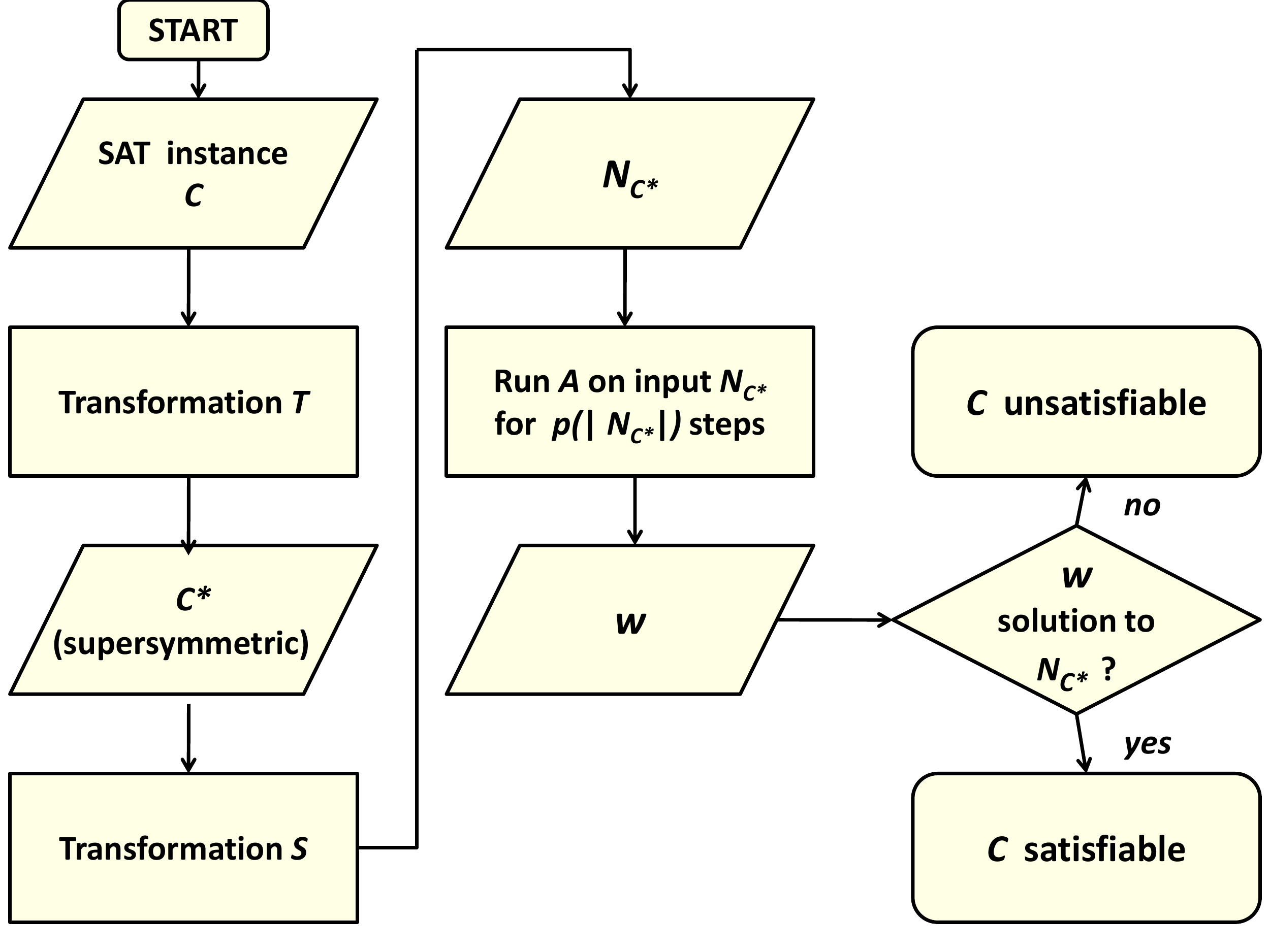}
\caption{Flowchart of the 3SAT-solver $A^*$}\label{fig:aijflowchart}
\end{figure*}

%\nop{
We are now ready for specifying our 3SAT-solver $A^*$ that works in polynomial time, and hence witnesses NP=P. Algorithm $A^*$ is also illustrated by the flowchart in Figure~\ref{fig:aijflowchart}.  The input of $A^*$ is  a 3SAT input instance $C$. We here assume without loss of generality that $C$ has at least two clauses. $A^*$ works as follows:
%} 

%Our polynomial-time 3SAT-solver $A^*$  is given below;  see also  the flowchart in %Figure~\ref{fig:aijflowchart}.  The input of $A^*$ is  a 3SAT input instance $C$. We here %assume without loss of generality that $C$ has at least two clauses. $A^*$ works as follows: 

\begin{enumerate}
\item Apply transformation $T$ to $C$ and get $C^*:=T(C)$. \\
{\em Note: $C^*$ is supersymmetric and $C^*$ is satisfiable iff $C$ is.}

\item Apply transformation $S$ to $C^*$ and get $N_{C^*}:= S(C^*)$.\\
{\em Note: $N_{C^*}$ solvable $\Leftrightarrow$  $C^*$ satisfiable $\Leftrightarrow$ $C$ satisfiable. }

\item \label{step3}Run $A$ on input $N_{C^*}$ for $p(|N_{C^*}|)$
steps; denote by $w$ the output at this point.\\
{\em Note:  If $C$  (and thus $C^*$)  is  satisfiable, then $N_{C^*}$ is a solvable minimal network, and thus $w$ is a solution to $N_{C^*}$; otherwise  
%$C^*$ and $C$ are  not satisfiable, and
 $N_{C^*}$ is unsolvable, and $w$ is the empty string or any string other than a solution to $N_{C^*}$.} 

\item Check if $w$ is a solution to $N_{C^*}$.

\item If $w$ is not a solution to $N_{C^*}$ then  output ``$C$ unsatisfiable" and stop. \\
{\em Note: In fact, 
if $w$ is not a solution to $N_{C^*}$  then $N_{C^*}$  is either empty or non-minimal. By the contrapositive of Fact $(*)$,  $C$ must then be unsatisfiable.}

\item  If $w$ is a solution to $N_{C^*}$ then output ``C satisfiable" and stop. \\
{\em Note: If $w$ is a solution, then $N_{C^*}$ is solvable, and thus $C^*$ and $C$ are satisfiable.}
\end{enumerate}

Each step of  $A^*$ requires  polynomial time only. The polynomial runtime of step~\ref{step3} depends parametrically  on the 
{\em fixed} polynomial $p$. $A^*$ is thus a polynomial-time 3SAT solver. The theorem for $k=2$ follows.

\smallskip

\smallskip 

Note that  $C^*$, as constructed in the proof of Theorem~\ref{theo:main}, is a 9SAT instance, hence the cardinality of the domain of 
each variable of $N_{C^*}$ is bounded by 9.

\smallskip

For $k>2$, the proof is analogous,
the main change being that  the transformation $S$  now creates an $\ell$-ary constraint  $c_L$ for each (ordered) set $L$ of $\ell\leq k$ clauses from $C$. The resulting constraint network $N_{C^*}=S(C^*)$, where $C^*$ is as constructed in Lemma~\ref{lem:sym} then does the job. 
\,  $\Box$

\subsection{The case of bounded domains}\label{subsec:bounded}

Theorem~\ref{theo:main} states that the problem of computing a solution from a non-empty minimal 
binary network is intractable even in case the cardinalities of the domains of all variables are bounded by a 
constant. However, if we take the total domain $dom(N)$, which is  the set of {\em all} literals of $C^*$, its cardinality is unbounded. 
This notwithstanding, the following simple corolloary to 
Theorem~\ref{theo:main} shows that even in case $|dom(N)|$ is bounded, computing a single solution from a minimal 
network $N$ is hard. 

\begin{corollary} \label{coroll:restricted} For each fixed $k\geq 2$, unless NP=P,  computing a single solution from a 
minimal $k$-ary constraint network $N$
 cannot be done in polynomial time, even in case 
$|dom(N)|$ is bounded by a constant.  
\end{corollary}

\noindent{\bf Proof.} \ 
We prove the result for $k=2$; for higher values of $k$, the proof is totally analogous.
The key fact we exploit here is that each variable $K_a$ of $N_{C^*}$ in the proof of Theorem~\ref{theo:main}
has a domain of exactly nine elements, corresponding to the nine literals occurring in clause $K_a$ of $C^*$.  We  
``standardize" these domains by simply renaming the nine literals  for each variable  by the numbers $1$ to $9$. We thus get an equivalent minimal constraint network with a total domain of cardinality 9. 
\nop{Thus for each $K_a$ we have a bijection $f_a: dom(K_a) \longleftrightarrow$ $\{1,2,\ldots,9\}$. 
Of course the same literal $\ell$ may be represented by different numbers for different variable-domains, i.e., 
it may well happen that  $f_a(\ell)\neq f_b(\ell)$. Similarly,  a value $i$ for $X_a$ may correspond to a completely different literal than 
the same number $i$ for $X_b$,  i.e., $f_a^{-1}(i)$ may well differ from $f_b^{-1}(i)$. Let us thus simply translate $N_{C^*}$ 
into a network $N^\#_{C^*}$, where each literal $\ell$ in each column of a variable $X_a$ is replaced by $f_a(\ell)$. It is easy to see that 
$N_{C^*}$  and $N^\#_{C^*}$  are equivalent and that the solutions of $N_{C^*}$ and   $N^\#_{C^*}$ are in a 
one-to-one relationship. Obviously,  $N^\#_{C^*}$ inherits from $N_{C^*}$ the property to be minimal in case it is solvable. 
Therefore, computing a solution to a network in which only nine values occur in total in the constraint relations is 
intractable unless NP=P.\,} $\Box$

\section{Minimal Network Recognition and Structure Identification}~\label{sec:ident}

\nop{In this section we consider a number of decision problems related to minimal networks that have 
been posed in the literature. First we}
In this section we first deal with the complexity of recognizing whether a  $k$-ary network $M$ is 
the minimal network of a $k$-ary network $N$ (Section~\ref{subsec:id1}).  We then study the problem 
of deciding whether a $k$-ary network $M$ is the minimal network of a single constraint (Section~\ref{subsec:id2}). 

\subsection{Minimal network recognition}~\label{subsec:id1}

An algorithmic problem of obvious relevance is recognizing whether a given network is minimal. Using the graph-theoretic characterization 
of minimal networks described in Section~\ref{sec:def}, Gaur~\cite{Gaur95} has
shown the following for binary networks:  

\begin{proposition}[Gaur~\cite{Gaur95}]
Deciding whether a complete binary network $N$ is minimal is NP-complete under Turing reductions.  
\end{proposition}

We generalize Gaur's result to the $k$-ary case and slightly strengthen it by showing NP-completeness under the standard notion of polynomial-time many-one reductions:

\begin{theorem}~\label{theo:recogn}
For each $k\geq 2$, deciding whether a complete $k$-ary network $N$ is minimal is NP-complete, even in case of bounded domain sizes.  
\end{theorem}

\noindent{\bf Proof.}\ 
Membership in NP is easily seen: We just need to guess a candidate
solution $s_t$ from $sol(N)$ for each of the polynomially many tuples
$t$ of each constraint $c$ of $N$, and check in polynomial time that $s_t$ is effectively a
solution and that the projection of $s_t$ over $scope(c)$ yields $t$. 
For proving hardness, revisit the proof of Theorem~\ref{theo:main}\nop{and consider its generalization, Theorem~\ref{theo:genrestricted} whose proof  works just in the same way (see~\cite{GG-CORR})}. 
For each $k\geq 2$, from a 3SAT instance $C$, we there construct in polynomial time a 
highly symmetric $k$-ary network with bounded domain sizes $N_{C^*}$, such that $N_{C^*}$ is minimal (i.e., $M_k(N_{C^*})=N_{C^*}$ iff  $C$ is satisfiable). This is clearly a 
standard many-one reduction from 3SAT to network minimality. 
\nop{NP-hardness follows.}    
\, $\Box$

A result in database theory similar to Theorem~\ref{theo:recogn} was shown in~\cite{Honeyman}, where it was proven that 
determining whether a set of database relations is join consistent (i.e., admits a universal relation)  is NP-complete. This was actually  proven for sets of binary relations, however not over schema $S_k$. Here we  showed  that this also holds for {\em complete $k$-ary  networks}, i.e., for sets of relations ove the specific schemas $S_k$, for each $k\geq 2$.  

\subsection{Structure identification and $k$-representability}~\label{subsec:id2}
\nop{So far we have looked at minimal networks $M_k(N)$ for $k$-ary
networks $N$. However, even for arbitrary networks a representation as 
an equivalent minimal network may be profitable, in particular, for 
networks consisting of a single constraint.} This section 
as well as Sections~\ref{bi-valued} and~\ref{frontier} 
 are dedicated 
to the problem of representing single constraints 
(or single-constraint networks) through equivalent $k$-ary minimal networks. By a slight abuse of terminology, when there is no danger of confusion, we will often
identify a single-constraint network $\{\rho\}$ with its unique
constraint $\rho$, and for tuples $t$ of the relation $rel(\rho)$ of the constraint $\rho$, we may write $t\in \rho$ instead of $t\in rel(\rho)$.

\nop{, and will thus often speak about a {\em single-constraint}
network $\rho$ when it actually would be correct to say $\{\rho\}$, and vice versa.}

\begin{definition}~\label{defn:min}
A complete $k$-ary network $M$ 
is a {\em minimal $k$-ary network} of $\rho$ iff 
\begin{enumerate}
\item $sol(M)=\rho$, and
\item  every tuple occurring in some constraint $r$ of $M$ is the projection of some tuple $t$ of $\rho$ over $scope(r)$. 
\end{enumerate}
\end{definition}

We say that a constraint relation $\rho$ is {\em $k$-representable}
if there exists a (not necessarily complete) $k$-ary constraint 
network $M$ such that $sol(M)=\rho$. 
The following proposition seems to be well-known and follows very easily 
from Definition~\ref{defn:min} anyway. 

\begin{proposition}~\label{prop:triv} Let $\rho$ be a constraint. The following three statements are equivalent: 
\begin{enumerate}
\item $\rho$ has a minimal $k$-ary network; 
\item $sol(\Pi_{S_k}(\rho))=\rho$;
\item $\rho$ is $k$-representable.
\end{enumerate}
\end{proposition}

Note that the equivalence of $\rho$ being $k$-representable and 
of $\rho$ admitting a minimal $k$-ary network 
emphasizes the importance and usefulness 
of minimal networks. In a sense this
equivalence means that the minimal $k$-ary network of $\rho$, if it exists, 
already represents  all $k$-ary networks that are equivalent to $\rho$. 

The complexity of deciding whether a minimal $k$-ary network for a relation
$\rho$ exists has been stated as an open problem by Dechter and Pearl 
in~\cite{dech-pear-92}. More precisely, Dechter and Pearl 
consider the equivalent problem of deciding whether  
$sol(\Pi_{S_k}(\rho))=\rho$ holds, and refer to this problem as a
problem of 
{\em structure identification in relational data}~\cite{dech-pear-92}. The idea 
is to identify the class of relations  $\rho$ that have the structural
property of being equivalent to the  $k$-ary  
network  $\Pi_{S_k}(\rho)$, and thus, of being $k$-representable. 
Dechter and Pearl formulated the following
conjecture: 

\begin{conj}[Dechter and Pearl~\cite{dech-pear-92}]~\label{conj2}
For each fixed positive integer $k\geq 2$, deciding whether 
$sol(\Pi_{S_k}(\rho))=\rho$ is NP-hard\footnote{Actually, 
the conjecture stated in~\cite{dech-pear-92} is somewhat weaker: 
Given a relation $\rho$ and an integer $k$,  deciding whether 
$sol(\Pi_{S_k}(\rho))=\rho$ is NP-hard. Thus $k$ is 
not fixed and is part of the input instance. 
However, from the context and use of this conjecture in~\cite{dech-pear-92} it is clear that 
Dechter and Pearl actually intend NP-hardness for each fixed $k\geq 2$.}. 
\end{conj}

As already observed by Dechter and Pearl  in~\cite{dech-pear-92}, there 
is a close relationship between the $k$-representability of constraint relations and a relevant database problem. Let us briefly 
digress on this. It is common 
knowledge that a single constraint $\rho$    
can be identified with 
a {\em data relation} in the context of relational databases (cf.~\cite{Dech03}). The decomposition of relations  plays an important role in the database area, in particular in the context of normalization~\cite{maier-83}. It consists of decomposing a relation $\rho$ without loss of information 
into smaller relations whose natural join yields precisely $\rho$. 
\nop{Decomposition can be conceived 
at the {\em instance level}, when a concrete  
relational instance is given, or at the {\em schema level} when 
a number of data dependencies are given with a schema. In our case, 
decomposition at the instance level is relevant.} If $\rho$ is a concrete data relation (i.e., a relational instance), and $S$ is a family of subsets (subschemas) of the schema of $\rho$, then the decomposition of $\rho$ over $S$ consists of the projection $\Pi_S=\{\Pi_s(\rho)\mid s\in S\}$ of $\rho$ over all schemes in $S$. This decomposition is {\em lossless} iff 
the natural join of all $\Pi_s(\rho)$ yields precisely $\rho$, or, equivalently, iff $\rho$ satisfies the {\em join dependency} $*[S]$.
We can thus reformulate the concept of $k$-decomposability in terms of database theory as follows: A relation $\rho$ is $k$-decomposable iff 
it satisfies the join dependency $*[S_k]$, i.e., iff the 
decomposition of $\rho$ into schema $S_k$ is lossless. The following 
complexity result was shown as early as 1981 in 
~\cite{MSY81}\footnote{
As mentioned by
Dechter and Pearl~\cite{dech-pear-92}, 
Jeff Ullman has proved this result, too. 
In fact, Ullman, on a request by Judea Pearl, 
while not aware of the specific 
result in~\cite{MSY81}, has produced a totally independent 
proof in 1991, and sent it as a private communication to 
Pearl. The result is also implicit in Moshe Vardi's 
1981 PhD thesis.} 

\begin{proposition}[Maier, Sagiv, and Yannakakis~\cite{MSY81}]~\label{prop:folk} 
Given a relation $\rho$ and a family $S$ of subsets of the schema of 
$\rho$, it is coNP-complete to determine 
whether $\rho$ satisfies 
the join dependency $*[S]$, or equivalently, whether the decomposition of $\rho$ into schema $S$ is lossless.
\end{proposition} 

Proposition~\ref{prop:folk} is weaker than  
Conjecture~\ref{conj2} and does not by itself imply it, 
nor so does its proof given in~\cite{MSY81}. In fact, Conjecture~\ref{conj2} speaks about the very specific sets $S_k$ for $k\geq 2$, which are neither mentioned in Proposition~\ref{prop:folk}
nor used in its proof. Actually, the NP-hardness proof in~\cite{MSY81} transforms 3SAT into 
the problem of checking a join dependency $*[S]$ over schema $S=(S_1,\ldots,S_{m+1})$, 
where one of the relation schemas, namely $S_{m+1}$ is of unbounded arity (depending on the input 3SAT instance), while the others are of arity 4.   
To prove Conjecture~\ref{conj2}, that refers to the specific schema $S_k$ in which all relations have arity at most $k$, we thus needed to develop a new and independent hardness argument.
 
\normalfont\begin{theorem}~\label{ident}
For each fixed integer $k\geq 2$,  deciding 
for a single constraint  $\rho$
whether 
$sol(\Pi_{S_k}(\rho))=\rho$, that is,  whether $\rho$ is 
$k$-decomposable, is coNP-complete. 
\end{theorem}

\noindent{\bf Proof.}
We show that deciding whether $sol(\Pi_{S_k}(\rho))\neq\rho$
is NP-complete. 

{\it Membership.} Membership in NP already follows from 
Proposition~\ref{prop:folk}, but we give a short proof of it here for sake of self-containment.
Clearly, $\rho \subseteq sol(\Pi_{S_k}(\rho))$. Thus $sol(\Pi_{S_k}(\rho))\neq\rho$ iff the containment is proper, which means that there 
exists a tuple $t_0$ in $sol(\Pi_{S_k}(\rho))$ not contained in $\rho$. One can guess such a 
tuple $t_0$ in polynomial time and check in polynomial time that for each $\ell\leq k$, each $\ell$-tuple of variables $X_{i_1},\ldots, X_{i_\ell}$ of  $var(\rho)$,  the projection of $t_0$ to 
$(X_{i_1},\ldots, X_{i_\ell})$ is indeed a 
tuple of the corresponding constraint of $S_k$. Thus determining whether $sol(\Pi_{S_k}(\rho))\neq\rho$ is in NP.
 
{\it Hardness.} We first show hardness for the binary case, that is, the case where  $k=2$. We use the NP-hard problem 3COL of deciding whether a graph $G=(V,E)$ with set of vertices $V=\{v_1,\ldots,v_n\}$ and edge 
set $E$ is 3-colorable. Let $G$ be given as input instance. 
We assume without loss of generality that $G$ has at least three vertices. Let $r$, $g$, $b$ be three data values standing  
for the three colors red, green, and blue, respectively. 
Let $\ncol$ be the constraint network defined as follows. The set of variables 
$var(\ncol)=\{X_1,\ldots,X_n\}$. 
The schema $S_2^+$ of $\ncol$ consists of all exactly binary scopes $(X_i,X_j)$ where $X_i\prec X_j$, and $dom(X_i)=\{r,g,b\}$ for $1\leq i\leq n$. Moreover,
for all $1\leq i<j\leq n$, the constraint $c_{ij}$ with schema
$(X_i,X_j)$ has the following constraint relation $rel(c_{ij})=r_{ij}$: 
if $(i,j)\in E$, then $r_{ij}$ is the set of pairs representing all legal vertex colorings, 
i.e., $r_{ij}=\{(r,g), (g,r), (r,b), (b,r), (g,b),(b,g)\}$; 
otherwise $r_{ij}=\{r,g,b\}^2$. 
$\ncol$ is thus a 
straightforward encoding of 3COL over schema $S_2^+$, and obviously $G$ is 3-colorable 
iff $sol(\ncol)\neq\emptyset$. Thus, deciding whether 
$sol(\ncol)\neq\emptyset$ is NP-hard.

We construct from $\ncol$ a single  constraint  $\rho$ 
with schema $\{X_1,\ldots,X_n\}$ 
as follows. The domain $dom(\rho)$ contains the color constants $r$, $g$, and $b$, as well as special "tuple identifiers" to be detailed below.  For each constraint $c_{ij}$ of $\ncol$, and for each tuple $(a,b)\in r_{ij}$,  $\rho$ contains a tuple $t$ whose $X_i$ and $X_j$ values are $a$ and $b$, respectively, and whose $X_\ell$ value, for all 
$1\leq \ell\leq n$, $\ell\neq i$, $\ell\neq j$, is a 
constant  $d^t_{ij}$, 
different from all values used in other tuples, whose purpose is to act as a tuple-identifier. 
This concludes the description of  the transformation from a 3COL instance $G=(V,E)$ to a constraint network $\ncol$ and further to a constraint  $\rho$. Clearly, this 
transformation is feasible in polynomial time. 
We claim the following:

\smallskip

\noindent CLAIM: $sol(\Pi_{S_2}(\rho))\neq\rho$ iff 
$sol(\ncol)\neq\emptyset$ (and thus iff $G$ is 3-colorable). 

\smallskip

This claim clearly 
implies the NP-hardness of deciding $sol(\Pi_{S_k}(\rho))\neq\rho$.
Let us prove that the claim holds. 

We start with the {\em if} direction.
Assume  $sol(\ncol)\neq\emptyset$. Then $G=(V,E)$ is 3-colorable and hence there exists a 
function $f:V\longrightarrow \{r,g,b\}$ such that for each edge $\langle v_i,v_j\rangle\in E$, 
$f(v_i)\neq f(v_j)$. Let $t$ be the tuple defined by: $\forall 1\leq i\leq n, t[X_i]=f(v_i)$. 
Then, by definition of $t$ and $\rho$, for each $1\leq i<j \leq n$, 
$t[X_i,X_j]\in \Pi_{X_i,X_j}(\rho)$ and fort each $1\leq i\leq n$, 
$t[X_i]\in \Pi_{X_i}(\rho)$. 
Therefore, $t\in sol(\Pi_{S_2}(\rho))$. However, $t\not\in \rho$, because each tuple of $\rho$, unlike $t$, has some tuple identifiers as components. It thus follows that $sol(\Pi_{S_2}(\rho))\neq\rho$.

Let us now show the {\em only if} direction of the claim.
Assume $sol(\Pi_{S_2}(\rho))\neq\rho$. 
Given that, as already noted, $\rho\subseteq  sol(\Pi_{S_2}(\rho))$, there must exist a tuple $t_0\in  sol(\Pi_{S_2}(\rho))$
such that $t_0\not\in \rho$. We show that $t_0$
can contain values from $\{r,g,b\}$ only, and must, moreover, be a solution 
to $\ncol$. 
%----------------------------------------
\nop{First, assume that $t_0$ contains two
distinct  tuple identifiers, say, $d$, in column $X_a$ and
$d'$ in column $X_b$ with $a<b$. This would mean that
$\Pi_{X_aX_b}(\rho)$ contains the tuple 
$(d,d')$, and
thus, that $\rho$ itself contains a tuple with the two {\em distinct} 
tuple identifiers $d$, and $d'$, which clearly
contradicts the definition of $\rho$. It follows that at most one 
tuple identifier $d$ may occur in $t_0$.}
%-----------------------------------------------------------------------
Assume a tuple identifier $d=d^t_{ij}$ occurs as a component of $t_0$. By construction of $\rho$,  
$d$ occurs in precisely one single tuple
$t$ of $\rho$. 
\nop{Let $\ell$ be a component of $t$ with $t[X_\ell]=d$. Then, for each 
$1\leq a\leq n$, $a\neq \ell$, there is a binary relation $q_a$  with schema $\{X_\ell, X_a\}$ in
$\Pi_{S_2}(\rho)$ such that  $q_a$ contains a tuple one of whose components is $d$ and the 
other one $t[X_a]$. Moreover, since $d$ occurs in a single tuple of $\rho$ only, each relation of 
$\Pi_{S_2}(\rho)$ has at most one tuple containing $d$.}
It follows that each relation of 
$\Pi_{S_2}(\rho)$ has at most one tuple containing $d$, and 
therefore the join of all relations of 
$\Pi_{S_2}(\rho)$ contains a single tuple only in which $d$ occurs as data value, namely $t$ itself.
\nop{It follows that $t$ is the only tuple of $sol(\Pi_{S_2}(\rho))$ containing 
$d=d^t_{ij}$ as a data value.} Therefore, $t_0=t$, and hence $t_0\in\rho$, which 
contradicts our assumption that $t_0\not\in \rho$.
We have thus shown that  $t_0$ cannot contain any tuple identifier at all, and can 
be made of ``color elements" from $\{r,g,b\}$ only. However,
by definition of $\rho$, 
each tuple $t_{ij}\in\{r,g,b\}^2$ occurring in a relation 
with schema $(X_i,X_j)$ of 
$\Pi_{S_2}(\rho)$ also occurs in the corresponding relation 
of $\ncol$, and vice versa. Thus 
$sol(\Pi_{S_2}(\rho))\neq\rho$ iff $sol(\ncol)\neq\emptyset$ iff 
$G$ is 3-colorable, which proves our claim.

For each fixed $k>2$ we can apply exactly the same line of reasoning. 
We define 
$\nkcol$ as the complete network on variables $\{X_1,\ldots,X_n\}$
of all $k$-ary correct  ``coloring''  constraints, where the relation 
with schema  $X_{i_1},\ldots,X_{i_k}$ expresses the correct colorings
of vertices $v_{i_1},\ldots,v_{i_k}$ of graph $G$.
We then define $\rho$ in a similar way as for $k=2$: each 
$k$-tuple of a relation of $\nkcol$ 
is extended by use of (possibly multiple occurrences of) a tuple identifier to an $n$-tuple of $\rho$. 
Given that $k$ is fixed, $\rho$ can be constructed in 
polynomial time, and so $\Pi_{S_k}(\rho)$.  
It is  readily seen that 
%two distinct tuple identifiers cannot jointly occur in a tuple of $sol(\Pi_{S_k}(\rho))$, and that 
each tuple of   
$sol(\Pi_{S_k}(\rho))$ that contains a tuple identifier is already present 
in $\rho$ because for each tuple-identifier value $d$, each 
relation of $\Pi_{S_k}(\rho)$ contains at most one tuple involving
$d$. Hence, any tuple in 
$sol(\Pi_{S_k}(\rho)) - \rho$ involves values from 
$\{r,g,b\}$ only, and is a solution to $\nkcol$ and thus a 
valid 3-coloring of $G$. 
\, $\Box$

\subsection{The case of bi-valued relations}\label{bi-valued}

Let us  now turn our attention to bi-valued relations $\rho$, that is, relations $\rho$ over a binary domain. 
As explained in Section 3.2 of~\cite{dech-pear-92}, such  bi-valued relations are of special interest, as that they correspond to Boolaen formulas. For example, a 3CNF can be seen as a bi-valued constraint network of ternary relations, and a single bi-valued relation $\rho$ corresponds to a DNF. The problem of structure identification in the bi-valued case thus corresponds to relevant identification and learnability questions about Boolean formulas; we refer the reader to~\cite{dech-pear-92} for details. In this context, it would be interesting to know whether, or for which parameter $k$,  Theorem~\ref{ident} carries over to the bi-valued case. 
While the coNP-membership clearly applies to the special case of a bi-valued $\rho$, the hardness part of that proof uses a multiple-valued relation $\rho$ and does not allow us to derive a hardness result for the bi-valued case. In fact,  the relations  $\rho$ constructed in the proof of Theorem~\ref{ident} from arbitrary 3COL instances are not bi-valued and actually have unbounded domains $dom(\rho)$ containing 
$|\rho|+3$ elements\footnote{Here $|\rho|=|rel(\rho)|$ designates the number of tuples in the constraint relation of the constraint $\rho$.}: 
the ``color constants" $r,g,b$, and the $|\rho|$ tuple identifiers $d^t_{i,j}$.

As noted in~\cite{Dech92,Gaur95}, for $k=2$ and bi-valued domains,
the problem of deciding whether $sol(\Pi_{S_k}(\rho))=\rho$ is tractable.  It can actually be reduced to 2SAT. But what about for values $k\geq 3$? Dechter and Pearl made the following 
conjecture (Conjecture 3.27 in~\cite{dech-pear-92}):

\begin{conj}[Dechter and Pearl~\cite{dech-pear-92}]~\label{conj3}
For each fixed positive integer $k\geq 3$, deciding for a bi-valued relation $\rho$ whether 
$sol(\Pi_{S_k}(\rho))=\rho$ is NP-hard.\footnote{Note that in~\cite{dech-pear-92}, the parameter  $k$ is not explicitly required to be fixed, however, from the context it is clear that the present stronger version of the conjecture was actually intended.  Moreover, Conjecture 3.27 in~\cite{dech-pear-92} was formulated in terms of  $k$-CNFs rather than in a purely relational setting. To avoid additional definitions and terminology, we have restated an equivalent relational formulation here. In particular,
we have replaced the term $M(\Gamma_{S_k}(\rho))$  in the original formulation by the  equivalent term  $sol(\Pi_{S_k}(\rho))$.} 
\end{conj}

We are able to confirm this conjecture.

\normalfont\begin{theorem}~\label{ident2}
For each fixed integer $k\geq 3$,  deciding 
for a single bi-valued constraint  $\rho$
whether 
$sol(\Pi_{S_k}(\rho))=\rho$, that is,  whether $\rho$ is 
$k$-decomposable, is coNP-complete. 
\end{theorem}

The rather involved proof of this theorem is given in~\ref{AppendixB}.
The  hardness part is similar  in spirit  to the one of  Theorem~\ref{ident}, except for two important changes that are due to the requirement of a two-valued domain. First we encode 3SAT rather than 3COL, in order to achieve a binary domain. However, there is still the problem of the tuple identifiers
(the values  $d^t_{i,j}$ in the proof of  Theorem~\ref{ident}). They are values from an unbounded domain.  We therefore use a specific bit-vector encoding that allows us to represent tuple identifiers in  binary format. This is, however, not totally trivial. The difficulty lies in the fact that in  the relations of the projection $\Pi_{S_k}(\rho)$ we do no longer  have full bit vectors at our disposal, but only $k$-bit projections of such bit vectors.  Sophisticated coding tricks are used for coping with this problem, and for obtaining a correct reduction.

\medskip

Theorem~\ref{ident2} has a  corollary, which we here formulate  in the terminology of Dechter and Pearl~\cite{dech-pear-92}. 

\begin{corollary}\label{idcol}
For fixed $k\geq 3$, the class of $k$-CNFs is not identifiable relative to all CNFs  (unless P=NP). 
\end{corollary}

The above  means the following. If a CNF  $\phi$  (or, more generally, a Boolean function $\phi$) is given by the set of all its models (i.e., by a bi-valued relation, each tuple of which corresponds to a model), then it is NP-hard to decide whether $\phi$ is equivalent to a $k$-CNF. We refer the reader to~\cite{dech-pear-92} for a more detailed account of $k$-CNF identification and its equivalence to the problem of whether a bi-valued relation $\rho$ is $k$-decomposable. To conclude this topic, let us note that the representation of a Boolean function $\phi$ by the explicit set of all its models, i.e., by all satisfying truth value assignments, is also known as  the {\em onset} of $\phi$~\cite{umans}.  The above  corollary thus states that, for fixed $k\geq 3$, it is NP-hard to decide whether a Boolean function specified  by its onset is equivalent to a $k$-CNF.\footnote{While we have not found this result in the literature on Boolean functions, we cannot totally exclude that it has been independently derived, maybe in a different context or using a different formalism.}

\subsection{Further strengthening  and tractability frontier}\label{frontier}

The  technique used to prove Theorem~\ref{ident2} can be used to strengthen Theorem~\ref{ident}, and to show  that it actually also holds 
for tri-valued constraints $\rho$.

\normalfont\begin{theorem}~\label{ident3}
For each fixed integer $k\geq 2$,  deciding 
for a single tri-valued constraint  $\rho$  
whether 
$sol(\Pi_{S_k}(\rho))=\rho$, that is,  whether $\rho$ is 
$k$-decomposable, is coNP-complete. 
\end{theorem}

The proof of this theorem is given in~\ref{AppendixC}. We  there use a transformation from 3COL from a graph $G=(V,E)$ as described in the proof of Theorem~\ref{ident} by applying, in addition,  similar vectorization techniques as in the proof of 
Theorem~\ref{ident2}.

\medskip

The above result, together with Theorem~\ref{ident2}, and with the fact that 
the $2$-representability of binary networks is feasible in polynomial time,  (see~\cite{Dech92}), and with the  facts that the $0$-representability and $1$-representability of each network 
and the $k$-representability of $1$-valued networks are 
trivially tractable, gives us the following precise characterization of the tractability of deciding whether $sol(\Pi_{S_k}(\rho))=\rho$:

\begin{theorem}\label{summary}
For the class of $i$-valued relations $\rho$, deciding  $sol(\Pi_{S_k}(\rho))=\rho$ is tractable iff  $i=1$ or ($i=2$ and $k\leq 2$). In all other cases, the problem is coNP-complete. 
\end{theorem}

The following figure illustrates this tractability frontier.

\smallskip
\begin{figure*}[h]
\centering
\includegraphics[width=0.7\textwidth,clip=true,trim=0cm 12.4cm 0cm 0cm]{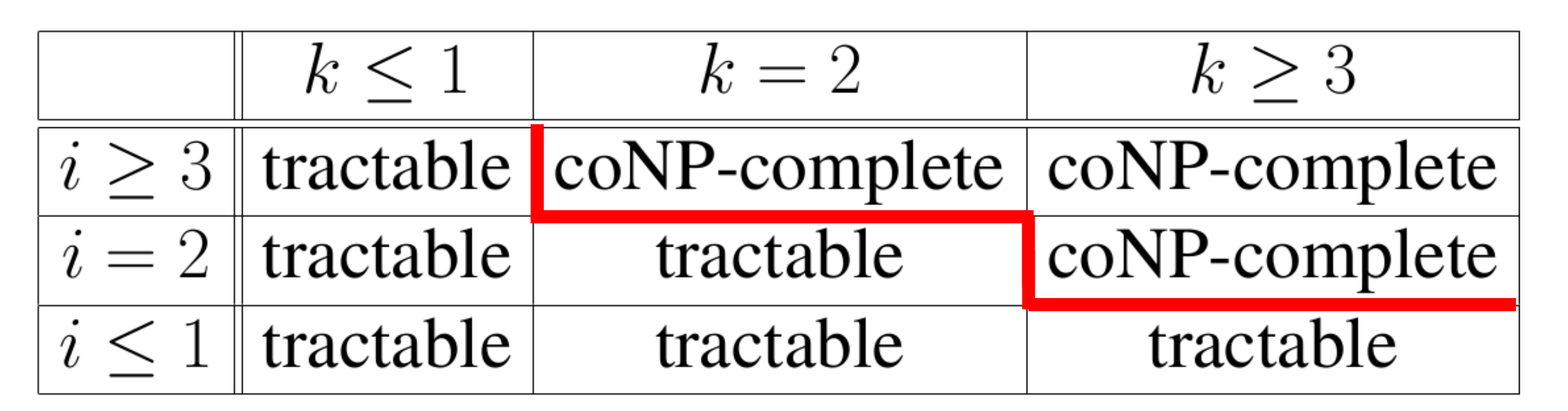}
\caption{Tractability Frontier for the $k$-decomposability  of $i$-valued relations $\rho$.}
\label{fig:tracTable}
\end{figure*}
\nop{
\bigskip 
\begin{center}
\begin{table}[h]
\centering
 \begin{tabular}{ |c || c | c | c | }
\hline 
%$i$-valued $\rho$:			
    &$k\leq 1$ & $k=2$ & $k\geq 3$ \\
\hline\hline  
  $i\geq 3$ & tractable &  coNP-complete & coNP-complete \\
  \hline $i=2$ & tractable  & tractable & coNP-complete \\
 \hline $i\leq 1$ & tractable & tractable & tractable \\
\hline  
\end{tabular}
\caption{Tractability Frontier for the $k$-decomposability  of $i$-valued relations $\rho$.}
\label{fig:tracTable}
\end{table}
\end{center}
} %end nop
\section{Summary, discussion, and future research}~\label{sec:conclusion}
In this paper we have tackled and solved long standing 
complexity problems related to minimal constraint networks:
\begin{itemize}   
\item We solved an open problem posed by Gaur~\cite{Gaur95} in 
1995, and later by Dechter~\cite{Dech03}, 
by proving Dechter's conjecture and showing that 
computing a solution to a minimal constraint network is NP-hard. 
\item We proved a conjecture on  structure identification in relational data made in 1992 by Dechter and Pearl~\cite{dech-pear-92}. In particular, we showed that for $k\geq 2$,
it is coNP-complete to decide whether for a single constraint (or data relation) $\rho$, 
$sol(\Pi_{S_k}(\rho))=\rho$, and thus whether $\rho$ is 
$k$-decomposable.
\item We also proved a refined conjecture of  Dechter and Pearl~\cite{dech-pear-92}, showing that the above problem remains coNP-hard  even if $\rho$ is a bi-valued constraint, in case $k\geq 3$. A consequence of this is the NP-hardness of  identifying  $k$-CNFs relative to the class of all CNFs (when represented by the explicit enumeration of their models).
\item We finally proved that deciding whether 
$sol(\Pi_{S_k}(\rho))=\rho$ is coNP-complete for tri-valued relations and $k\geq 2$. Together with our other results on structure identification, this allowed us to trace the precise tractability frontier for this problem.
\end{itemize}

We wish to make clear that our hardness result about 
computing solutions to minimal networks does not
mean that we think minimal networks are useless. To the contrary, we are convinced that network minimality is a most desirable property when a solution space needs to be efficiently represented for applications such as computer-supported configuration~\cite{FFetal98}. 
For example, a user interactively configuring a PC constrains a relatively small number of variables, say, 
by specifying a maximum price, a minimum CPU clock rate, 
and the desired hard disk type and capacity. The user then
wants to quickly know whether a solution exists, and if so, wants to see it. For a $k$-ary minimal 
constraint network,  the satisfiability of queries involving $k$ variables only can be decided in polynomial time. However, our Theorem~\ref{theo:main} states 
that, unless NP=P, in case the query is satisfiable, 
there is no way to witness the satisfiability by a complete solution (in our example, by exhibiting a completely
configured PC satisfying the user requests).

Our Theorem~\ref{theo:main} thus unveils a certain deficiency of minimal networks, namely, the failure of 
being able to exhibit full solutions. However, we have 
a strikingly simple proposal for  redressing this deficiency. Rather than just storing $\ell$-tuples 
(where $\ell\leq k$) in a $k$-ary minimal network
$M_k(N)$, we may store a full solution $t^+$ with each $\ell$-tuple, 
where $t^+$ coincides with $t$ on the $\ell$ variables of  $t$. Call this extended minimal network $M^+_k(N)$. Complexity-wise, $M^+_k(N)$ is not harder to obtain than $M_k(N)$. Moreover, in practical terms, given that the known 
algorithms for computing $M_k(N)$ from $N$ require to check 
for each $\ell$-tuple $t$ whether it occurs in some solution $t^+$, why not just memorize $t^+$ on the fly for each ``good'' tuple $t$? Note also that the size of $M^+_k(N)$ is still polynomial, and at most by a factor $|var(N)|$ larger than the size of $M_k(N)$. One may even go further and store not just a single solution, but the $K$ {\em best} solutions (according to some predefined preference ordering) whose values coincide with those of $t$ with each tuple $t$ of $M_k(N)$. This allows one to answer {\em top K queries} with at most $k$ variables in polynomial time, once  $M_k(N)$ has been compiled. An example would be: {\em show me the 5 cheapest laptops fulfilling  $\phi$}, where $\phi$  constrains $k$ variables only.

For practical applications it is not always optimal to consider the complete schemas $S_k$ as defined here. In the conference version~\cite{GGCP11} of this paper, $S_k$ was defined  to contain only all {\em exactly} $k$-ary relations over a given set of variables, rather than all  {\em at most} $k$-ary relations. It is easy to see that  the relations with scopes of fewer than $k$ variables are redundant and can indeed be omitted (they can always be obtained via projections from the exactly $k$-ary relations). The only reason why we used the complete schemas in the present journal version is that we wanted to use exactly the same definition as in the standard 
references~\cite{Mont74,Dech03}. However, yet more liberal definitions are possible. For example,  Lecoutre~\cite{Lec} defines a constraint network $N$ over an arbitrary schema $S$ to be minimal if $N=\Pi_S(sol(N))$. Clearly, all our complexity bounds carry over to this more liberal setting: the lower bounds are directly inhererited, as all instances in our settings are also instances of the more liberal setting, and the upper bounds are obtained by a trivial adaptation of the proofs of our existing upper bounds. 

An interesting problem for future research is the following. We may issue queries of the 
following form against $M^+_k(N)$: \texttt{SELECT A SOLUTION WHERE $\phi$}. 
Here $\phi$ is some Boolean combination on constraints on the variables of $N$. 
Queries, where $\phi$ is a simple 
combination of range restrictions on $k$ variables can be answered in polynomial time. 
But there are much more complicated queries that can be answered efficiently, for 
example, queries that involve aggregate functions and/or 
re-use of quantified variables. It would thus be nice and useful to 
identify very large classes of 
queries to $M^+_k(N)$ for which 
a single solution -- if it exists -- can be found in polynomial time.  
\nop{Consider queries of the 
following form against $M^+_k(N)$: \texttt{SELECT A SOLUTION WHERE $\phi$}, where $\phi$ is a Boolean combination on constraints on the  variables of $N$. If $\phi$ is just a Boolean combination of range restrictions for  
$k$ variables, such queries can be answered in polynomial time. But  more complicated queries can be answered efficiently, for 
example, queries involving aggregate functions and/or 
the re-use of quantified variables. It would be nice to 
identify large classes of 
queries to $M^+_k(N)$ for which 
a single solution -- if one exists -- can be found efficiently.}  

Another relevant research problem is related to the projection $\Pi_{S^*} (sol(N))$ of the solution $sol(N)$ of a (not necessarily binary) constraint network $N$ to a user-defined schema $S^*$, and to the further use of the schema $S^*$ for distributed constraint solving.  
The projection of the solution space to specific sets of variables is used in the context of system configuration, when a system is jointly configured by a number of engineers, each having access to a projection of the solution space only~\cite{Voronov}.
The problem of computing a solution  $\Pi_{S^*} (sol(N))$ is generally NP-hard, and remains NP-hard in many special cases, e.g. if $N$ is binary and, at the same time  $S^*=S_2$ (see Theorem~\ref{theo:main}). We would like to investigate relevant restrictions that make this problem tractable. For some restrictions, this is already known. For example, if $S^*$ has bounded hypertree width~\cite{GLS02,GLS01,Adler}, then this problem becomes tractable. If $S^*$ has bounded hinge width, then computing a solution can even be done in a backtrack-free manner, see Section~3 of~\cite{Gyssens-etal-94}. Note that bounded hinge width is a stronger restriction than bounded hypertree width; for a comparison of these and other hypergraph restrictions, see~\cite{GLS00}.
Other decompositions that lead to backtrack solution search are the {\em world-set decompositions}  discussed in~\cite{Olteanu1} and further generalized in~\cite{Olteanu2}. These decompositions are based on Cartesian products rather than on joins, therefore, computing solutions is easier than with project-join decompositions.

There is also the problem of  computing the desired projections without computing the possibly very large relation $sol(N)$, and, as a special case, computing the minimal constraint network $M(N)$ from a given network $N$. More formally, we would like to compute  $\Pi_{S^*} (sol(N))$
from $N$ in polynomial space as efficiently as possible, assuming the relations of $S^*$ are all of bounded arity. There are already  promising approaches to this problem in the literature. 
In~\cite{Gebser07,Gebser09}, conflict-driven answer set programming (ASP) techniques are used for this task.  
In~\cite{Voronov}, projections of $sol(N)$ are computed via a SAT solver, and it is shown that this method is feasible for large datasets stemming from the automotive industry. However, we expect that a structural analysis of the original network $N$ and of the desired projection schema $S^*$ could further help to speed up this computation.

\appendix

\section{Proof of the Symmetry Lemma}\label{AppendixA}

\newenvironment{xtheorem}[2][Theorem]{\begin{trivlist}
\item[\hskip \labelsep {\bfseries #1}\hskip \labelsep {\bfseries #2}]}{\end{trivlist}}

\newenvironment{xlemma}[2][Lemma]{\begin{trivlist}
\item[\hskip \labelsep {\bfseries #1}\hskip \labelsep {\bfseries #2}]}{\end{trivlist}}

\begin{xlemma}{\ref{lem:sym}}[Symmetry Lemma]
{\em For each fixed integer $k\geq 1$, there is a poly\-no\-mial-time transformation $T$ 
that transforms each 3SAT instance $C$  
into a $k$-supersymmetric instance $C^*$ such that $C$ is satisfiable iff $C^*$ is satisfiable.} 
\end{xlemma}

\begin{proof}  
We first prove the lemma for $k=2$.
Consider the given 3SAT instance $C$. Create for each propositional variable $p\in propvar(C)$ a set 
$\New(p)=\{p_1, p_2, p_3,p_4,p_5\}$ of fresh propositional variables. Let $Disj^+(p)$  be the set of all 
disjunctions of three distinct positive atoms from $\New(p)$ and let $Disj^-(p)$ be the set of all 
disjunctions of three distinct negative literals corresponding to atoms in $\New(p)$. Thus, for example 
$(p_2\vee p_4\vee p_5)\in Disj^+(p)$ and $(\bar{p_1}\vee \bar{p_4}\vee \bar{p_5})\in Disj^-(p)$. 
Note that $Disj^+(p)$ and $Disj^-(p)$ each have exactly ${5\choose 3}=10$ elements (we do not distinguish between 
syntactic variants of equivalent clauses containing the same literals).

Consider the following transformation $T$, which eliminates all original 
literals from $C$, yielding $C^*$: 

\begin{quote}
\noindent{\bf Function T:}\\  
\noindent BEGIN
\noindent $C':=C$.\\ 
\noindent WHILE   $propvar(C)\cap propvar(C')\neq\emptyset$\, DO\\
\hspace*{0.5cm}$\{$\,pick any $p\in propvar(C)\cap propvar(C')$; $C':=elim(C',p)$$\}$;\\
\noindent Output($C'$)\\
\noindent END.
\end{quote}

Here $elim(C',p)$ is obtained from $C'$ and $p$ as follows:\\ 

\noindent FOR each clause $K$ of $C'$ in which $p$ occurs positively or negatively DO
\begin{description}
\item{} BEGIN
\item{} let $\delta$ be the disjunction of all literals in $K$ different from $p$ and from $\neg p$;\footnote{An empty $\delta$ is equal to \emph{false}, and it is understood
that $\alpha \vee \mbox{\emph{false}}$ is simply $\alpha$.}
\item{} if $p$ occurs positively in $K$, replace $K$ in $C'$ by the conjunction  $\Gamma^+(K)$ of all 
clauses of the form $\alpha \vee \delta$, where $\alpha\in Disj^+(p)$;
\item{} if $p$ occurs negatively in $K$, replace $K$ in $C'$ by the conjunction  $\Gamma^-(K)$ of all 
clauses of the form $\alpha \vee \delta$, where $\alpha\in Disj^-(p)$;
\item END.
\end{description}
\nop{
replace each clause $K$ in which $p$ occurs positively and which contains a
disjunction $\delta$ of further literals by the conjunction $\Gamma^+(K)$ of 
all  clauses of the form $\alpha\vee \delta$ where $\alpha\in Disj^+(p)$;
\item replace each clause $K$ in which $p$ occurs negatively and which contains a 
disjunction $\delta$ of further literals by the conjunction $\Gamma^-(K)$ of 
all  clauses of the form $\alpha\vee \delta$ where $\alpha\in Disj^-(p)$;
\end{itemize}}

Let $C^*=T(C)$ be the final result of $T$. $C^*$ contains no original variable from $propvar(C)$.
Note that  $C^*$ can be computed in polynomial time from $C$. In fact,  note that every clause of three literals of $C$ gives rise to exactly ${5 \choose 3}^{\rm 3}=10^3=1000$ clauses  of 9 literals each in $C^*$.
While computing $C^*$ from $C$, we can thus replace each 3-literal clause of $C$ at once and independently by the corresponding  1000 clauses.  Similar direct replacements (but with fewer result clauses) are, of course,  possible for two-literal and one-literal clauses of $C$. Assuming appropriate data structures, the transformation from $C$ to $C^*$  can thus actually be done in linear time. 

\smallskip

We now need to prove (1) that $C^*$ is satisfiable iff $C$ is and (2) that $C^*$ 
is 2-supersymmetric.

\smallskip

\noindent{\em Fact 1: $C^*$ is satisfiable iff $C$ is.}\  
We will actually prove more than we need here. In fact, our proof of Fact~1 below also shows that a satisfying assignment to $C$ can be transformed into {\em many} satisfying assignments to $C^*$. We will use this, when we come to prove supersymmetry in Fact~2.\  We prove Fact~1 by showing that, when 
at each step of algorithm $T$, $C'$ is transformed into its next value $C''=elim(C',p)$, then $C'$ and $C''$ are satisfaction-equivalent. 
Fact~1 then follows by induction.  Assume $C'$ is satisfied via a truth value assignment $\tau'$. 
Then let $\tau''$ be any truth value assignment to the propositional variables of $C''$ with the following properties:
\begin{itemize} 
\item For each propositional variable $q$ of $C''$ different from $p$, $\tau''(q)=\tau'(q)$;  
\item if $\tau'(p)=\true$, then at least 3 of the variables in $\New(p)$ are set true by $\tau''$, and 
\item if $\tau'(p)=\false$, then at most two of the variables in $\New(p)$ are set true by $\tau''$ (and at least three are thus set false).
\end{itemize} 
By definition of $C''$, $\tau''$ must satisfy $C''$. In fact, assume first $\tau'(p)=\true$. Let $K$ be a clause of $C$ in which 
$p$ occurs positively.  Then, given that 
at least three variables in $\New(p)$ are set true by $\tau''$, each element of $Disj^+(p)$ must have at least one atom made true 
by $\tau''$, and thus each of the clauses of $\Gamma^+(K)$ of $C''$ evaluates to true via $\tau''$. All other clauses of $C''$ stem from clauses 
of $C'$ that were made true by literals corresponding to an atom $q$ different from $p$. But, by definition of $\tau$, these literals keep their 
truth values, and hence make the clauses true. In summary, all clauses of $C''$ are satisfied by $\tau''$. In a very similar way it is shown  
shown that  $\tau''$ satisfies $C''$ if, $\tau'(p)=\false$. \\  
Vice versa, assume some truth value assignment $\tau''$ satisfies $C''$. Then it is not hard to see that $C'$ must be  satisfied by the  truth value 
assignment $\tau'$ to $C'$ defined as follows: If a majority (i.e. 3 or more) of the five atoms in $\New(p)$ are made true via $\tau''$, then let $\tau'(p)=\true$, 
otherwise let $\tau'(p)=\false$; moreover,  for all propositional variables $q\not\in \New(p)$, let $\tau'(q)=\tau''(q)$. \\
To see that $\tau'$ satisfies $C'$, consider first the case that three or more of the propositional variables of $\New(p)$ are assigned {\it true} 
by $\tau''$. Note that all clauses of $C'$ that neither contain $p$ nor $\bar{p}$ are trivially satisfied by  $\tau'$, as $\tau'$ and $\tau''$ 
coincide on their atoms. Now let us consider any clause $K$ of $C'$ in which $p$ occurs positively. Then the only clauses that contain 
positive occurrences of elements of $\New(p)$ of $C''$ are the sets  $\Gamma^+(K)$. If $\tau''$ is such that it makes at least three 
of the five atoms in $\New(p)$ true, then any clause in $\Gamma^+(K)$ is made true by atoms of $\New(p)$. Thus when replacing these 
atoms by $p$ and assigning $p$ true, the resulting clause $K$ remains true. Now consider a clause $K=\bar{p}\vee\delta$ of $C'$ in which $p$ occurs negatively.
The only clauses containing negative  $\New(p)$-literals in $C''$ are, by definition of $C''$,  those in $\Gamma^-(K)$. Recall we assumed that that $\tau''$ 
satisfies at least three distinct atoms from $\New(p)$. Let three of these satisfied atoms be $p_i$, $p_j$, and $p_k$. By definition, 
$\Gamma^-(K)$ contains a clause of the form $\bar{p_i}\vee \bar{p_j}\vee\bar{p_k}\vee\delta$. Given that this clause is satisfied by $\tau''$, 
but $\tau''$ falsifies   $\bar{p_i}\vee \bar{p_j}\vee\bar{p_k}$,  $\delta$ is satisfied by $\tau''$, and since $\delta$ contains no 
$\New(p)$-literals, $\delta$ is also satisfied by $\tau'$. Therefore, $K=\bar{p}\vee\delta$ is satisfied by $\tau'$. This concludes the case 
where three or more of the propositional variables of $\New(p)$ are assigned {\it true} by $\tau''$. The case where  three or more of the propositional variables
 of $\New(p)$ are assigned {\it false} by $\tau''$ is completely symmetric, and can thus be settled in a totally similar way. $\Diamond$

\smallskip

\noindent {\em Fact 2: Proof that $C^*$ is 2-supersymmetric}.
Assume $C^*$ is satisfiable by some truth value assignment $\eta$. 
Then $C$ is satisfiable by some truth value assignment $\tau$, and thus $C^*$ is satisfiable 
by some truth value assignment $\tau^*$ constructed inductively as described in the proof of Fact 1.  Let us have a closer look at the inductive construction used to obtain $\tau^*$ in Fact~1. For any initially fixed  pair of propositional variables 
$p_i,q_j\in propvar(C^*)$, where   $1\leq i, j\leq 5$, the construction of $\tau^*$ gives us a very large degree of freedom for chosing $\tau^*$. Actually, the construction is so general, that it allows us to  
let $p_i,q_j$ take on any arbitrary truth value assignment among of the four possible joint truth value assignments. 
In fact,  however we choose the truth value assignments for two among the variables in  $\{p_1,\ldots,p_5,q_1,\ldots,q_5\}$, there is always enough flexibility 
for assigning the remaining variables in this set some truth values that ensure that the majority of variables has the truth value required by 
the proof of Fact~1 for representing the original truth value of $p$ via $\tau'$. (This holds even in case $p$ and $q$ are one and the same variable, and we thus 
want to force two elements from $\{p_1,\ldots, p_5\}$ to take on some truth values, see the second example below.)   
Let us give two examples that illustrate the two characteristic cases to consider. First, assume $p$ and $q$ are distinct and $\tau$ satisfies $p$ and falsifies $q$.   
We would like to construct, for example,  a truth value assignment $\tau^*$ that falsifies  $p_2$ and simultaneously satisfies $q_4$. In constructing $\tau^*$, the 
only requirements on $\New(p)$ and $\New(q)$ are that more than three variables from $\New(p)$ need to be satisfied by $\tau^*$, but no more than two from $\New(q)$ need to 
be satisfied by $\tau^*$. For instance, we may then set $\tau^*(p_1)=\tau^*(p_3)=\tau^*(p_4)=\tau^*(p_5)=\true$ and $\tau^*(p_2)=\false$ and $\tau^*(q_1)=\tau^*(q_2)=\tau^*(q_3)=\tau^*(q_5)=\false$
and $\tau^*(q_4)=\true$. This achieves the desired truth value assignment to $p_2$ and $q_4$. An extension to a full satisfying truth value assignment $\tau^*$ for $C^*$ is guaranteed. 
Now, as a second example, assume that $\tau(p)=\false$, but we would like $\tau(p_1)$ and $\tau(p_2)$ to be simultaneously 
true in a truth value assignment satisfying $C^*$. Note that in this case, the only requirement on $\New(p)$ in the construction of $\tau^*$ is 
that at most two atoms from $\New(p)$ must be  assigned $\true$. Here we have a 
single option only: set $\tau^*(p_1)=\tau^*(p_2)=\true$ and $\tau^*(p_3)=\tau^*(p_4)=\tau^*(p_5)=\false$. This option works perfectly, and assigns the desired truth values to 
$p_1$ and $p_2$.\  In summary, $C^*$ is $2$-supersymmetric.   $\Diamond$

\smallskip 
The proof for $k>2$ is totally analogous, except for the following modifications: 
\begin{itemize} 
\item Instead of creating for each propositional variable $p\in propvar(C)$ 
a set \\ 
$\New(p)=\{p_1,p_2,\ldots,p_5\}$ of five new variables, we now create 
a set \\ 
$\New(p)=\{p_1,p_2,\ldots,p_{2k+1}\}$ of  $2k+1$ new propositional variables. 
\item The set $Disj^+(p)$ is now defined as the set of all disjunctions of 
$k+1$ positive atoms from $\New(p)$. Similarly, $Disj^-(p)$ is now defined as 
the set of all disjunctions of $k+1$ negative literals obtained by negating 
atoms from $\New(p)$.
\item We replace the numbers 2 and 3 by $k$ and $k+1$, respectively.
\item We note that now each three-literal clause of $C$ is replaced no longer by 
${5 \choose 3}^{\rm 3}$ clauses but by ${2k+1 \choose k+1}^{\rm 3}$ clauses. 
\item We note that the resulting clause set $C^*$ is now a $3(k+1)$-SAT instance.
\end{itemize}
It is easy to see that the proofs of Fact~1 and Fact~2 above go through 
with these modifications.

Finally, let us recall that any 2-supersymmetric SAT instance is trivially 
also 1-supersymmetric, which settles the theorem for $k=1$. 
\end{proof}

\smallskip

\section{Proof of Theorem \ref{ident2} }~\label{AppendixB}

\vspace{-0.5cm}
\begin{xtheorem}{\ref{ident2}}
{\em For each fixed integer $k\geq 3$,  deciding 
for a single bi-valued constraint  $\rho$
whether 
$sol(\Pi_{S_k}(\rho))=\rho$, that is,  whether $\rho$ is 
$k$-decomposable, is coNP-complete.} 
\end{xtheorem}

\begin{proof}
It suffices to show coNP-hardness, as membership in coNP already follows from Theorem~\ref{ident}.  We first prove coNP-hardness for  the case $k=3$.   

Consider a  nonemptyt 3SAT instance $C=\{C_1,\ldots,C_m\}$ over a set $propvar(C)= \{p_1,\ldots, p_n\}$ of propositional variables, where each $C_i$ is a clause containing precisely 3 literals whose corresponding atoms are mutually distinct. 

Let us first define two numbers $r_0$ and $r$  from $C$, whose meaning and use will become clear later on. Let $r_0$ denote the number of  3-element sets  $\{p_a,p_b,p_c\}$  of  mutually distinct propositional variables $p_a,p_b,p_c\in propvar(C)$ that do not all three jointly appear in any clause of $C$. 
Note that $r_0\leq 8{n\choose 3}$. Let, moreover, $r=7m+r_0$. Clearly, $r$ is polynomially bounded in the size of $C$, as 
$r\leq 7m+ 8{n\choose 3}$.

We construct in polynomial time a bi-valued constraint $\rho$  
of $r$ elements, such that $sol(\Pi_{S_k}(\rho))\neq\rho$ iff $C$ is satisfiable. The scope $scope(\rho)$ of $\rho$ contains for each $p_i\in propvar(C)$ a list of $r+1$ variables $X_i^0,X_i^1,\ldots,X_i^r$. 
Intuitively, in each tuple $t$ of $rel(\rho)$, for each $1\leq i\leq n$, the values assigned to the variables   $X_i^0,X_i^1,\ldots,X_i^r$
either shall encode a truth value assignment to $p_i$, in which case all variables of this list will be assigned the same value, zero or one,  or these values shall encode a tuple identifier for the tuple in which they occur.  A tuple identifier for the $s$-th tuple of $rel(\rho)$  
assigns the value zero to all $X_i^j$ where $j\leq s$ and the value one to all $X_i^j$ where $j\geq s$. This will be made more formal below. 

The constraint relation $rel(\rho)$ consists of two groups of tuples: 

\begin{description}
\item[Clause-induced tuples.] These are 
$7m$ tuples, namely, seven for each clause $C_h$, $1\leq h\leq m$.
These tuples are numbered from $1$ to $7m$. 
Each of these tuples describes one of the 7 legal truth value assignments (out of 8 possible) to the three propositional variables of a clause $C_h\in C$.   For each clause $C_h$, $1\leq h\leq m$, and each truth value assignment $\tau_j\in  propvar(C_h)\longrightarrow\{0,1\}$, among all 7 permitted truth value assignments to the propositional variables of $C_h$, where $1\leq j\leq 7$,  $rel(\rho)$ contains precisely one tuple 
$t_h^j$, whose components are described as follows.  For each $p_i\in propvar(C_h)$, 
$t_h^j[X_i^0]=t_h^j[X_i^1]=t_h^j[X_i^2]=\cdots = t_h^j[X_i^r]=\tau_j(p_i)$. Moreover, for each 
$p_i\in propvar(C)-propvar(C_h)$, $t_h^j[X_i^0]=0$, and the assignments to $X_i^1\ldots X_i^r$ 
jointly constitute a unique tuple identifier that exclusively appears in the tuple $t_h^j$, and that encodes the tuple number $s$ of the tuple $t_h^j$\, (namely $s=7(h-1)+j$)\, in a very simple way: It assigns $0$ to all $X_i^{s'}$ where $0\leq s'<s$ and  $1$ 
to all  variables $X_i^{s'}$ where $s\leq s'\leq r$.

\item[Auxiliary tuples]  These are no more than $8{n\choose 3}$ tuples: one for each 3-element set  $\{p_a,p_b,p_c\}$  of  mutually distinct propositional variables $p_a,p_b,p_c\in propvar(C)$ that do not all three jointly appear in any clause of $C$.  These auxiliary tuples are numbered from $7m+1$ to $r$, where 
$r\leq 7m+8{n\choose 3}$ is  the total number of tuples in $\rho$. Essentially, the eight auxiliary tuples associated with the above sets $\{p_a,p_b,p_c\}$ each encode one of the eight truth value assignments $\sigma_1,\ldots,\sigma_8$ to the propositional variables $p_a$, $p_b$, and $p_c$. These tuples thus do not encode effective constraints, as they reflect any arbitrary truth value assignment $p_a$, $p_b$, and $p_c$, but they will be needed for technical reasons. More formally, for each set $S=\{p_a,p_b,p_c\}$ as above, and each truth value assignment $\sigma$ to 
$\{p_a,p_b,p_c\}$, $rel(\rho)$ contains a tuple $t_S^\sigma$, whose components are described as follows.  For each $p_i\in S$, $t_S^\sigma[X_i^0]=t_S^\sigma[X_i^1]=t_s^\sigma[X_i^2]=\cdots = t_S^\sigma[X_i^r]=\sigma(p_i)$.
Moreover, for each 
$p_i\in propvar(C)-S$, $t_S^\sigma[X_i^0]=0$, and the assignments to $X_i^1\ldots X_i^r$, just as before, 
jointly constitute a unique tuple-identifier that exclusively appears in the tuple $t_S^\sigma$, and that encodes the tuple number $s$ of the tuple $t_S^\sigma$ by assigning $0$ to all $X_i^{s'}$ where $s'<s$ and  $1$ 
to all  variables $X_i^{s'}$ where $s'\geq s$. 
 \end{description}

This concludes the definition of $\rho$.

\medskip

\noindent 
CLAIM: $sol(\Pi_{S_3}(\rho))\neq\rho$ iff $C$ is satisfiable. 

\smallskip

We first prove the {\em if-part} of the claim. 
Assume $C$ is satisfiable. Thus  there exists a truth value assignment $\tau$ to $propvar(C)$ satisfying $C$. We show that  $sol(\Pi_{S_3}(\rho))$ then must contain the tuple $t\not\in rel(\rho)$
defined as follows.  For each $1\leq i\leq n$, $t[X_i^0]=t[X_i^1]=t[X_i^2]=\cdots = t[X_i^r]=\tau(p_i)$.  To see this, it suffices to observe that the projection $t[S]$ of $t$ to any set   $S=\{X_a^u,X_b^v,X_c^w\}$ of three distinct variables
from $scope(\rho)$ is  contained in the corresponding relation $\Pi_S(\rho)$ of $\Pi_{S_3}(\rho)$.  

In fact, if the atoms $p_a, p_b$ and $p_c$ jointly occur in a clause $C_h$ of $C$, then,  the tuple $t'$ in $rel(\rho)$ induced by $C_h$ for truth value assignment  $\tau[p_a, p_b,p_c]$
coincides in its  $S$-components with the tuple $t$, in other terms, $t'[S]=t[S]$. Hence $t[S]$ is contained in the relation $\Pi_S(\rho)$ of $\Pi_{S_3}(\rho)$. Moreover, in case  $p_a, p_b$ and $p_c$ do not jointly occur in a clause of $C$, then there must exist an auxiliary tuple $t'$ such that $t'[S]=t[S]$, and thus, again,  $t[S]$ is contained in the relation $\Pi_S(\rho)$ of $\Pi_{S_3}(\rho)$. In summary, $t$ is contained in the join of the exactly ternary relations of 
$\Pi_{S_3}(\rho)$. As is easily verified, the binary and unary relations of $\Pi_{S_3}(\rho)$ are weaker than the ternary ones, and actually redundant; the join of all constraints with precisely three variables is in fact equal to $sol(\Pi_{S_3}(\rho))$. 
It follows that $t$ is contained in the join of $\Pi_{S_3}(\rho)$, which is $sol(\Pi_{S_3}(\rho))$. However, $t$ is not in $rel(\rho)$ because $t$ does not contain any tuple identifier, whereas each tuple of $rel(\rho)$ does.

It now remains to show that,  whenever $sol(\Pi_{S_3}(\rho))$ contains a tuple $t\not\in rel(\rho)$, then $t$ corresponds to a satisfying truth value assignment for $C$, and $C$ is thus satisfiable. Let $t$ be such a tuple. 
We first show that for each $1\leq i\leq n$ and each $1\leq v\leq r$ and $1\leq w\leq r$ it must hold that $t[X_i^v]=t[X_i^w]$, thus all bits of $t[X_i^0,X_i^1,\ldots,X_i^r]$ must be  equal. We prove this by showing that this bit-vector cannot have two consecutive bits of different value.
\begin{itemize}
\item Assume that for some $0< \ell \leq r$, $t[X_i^{\ell-1}]=0$ while  $t[X_i^{\ell}]=1$.  By construction,  
$rel(\rho)$ contains only a single tuple $t'$ for which $t'[X_i^{\ell-1}]=0$ but  $t'[X_i^{\ell}]=1$, namely the tuple numbered $\ell$.
Therefore, in each relation $rel(c)$ of any constraint $c$  of $\Pi_{S_3}(\rho)$ where $scope(c)$ contains $X_i^{\ell-1},X_i^{\ell}$ and any other variable $X_j^u$, there is thus a single tuple $f_c$ having $f_c[X_i^{\ell-1}]=0$ and $f_c[X_i^{\ell}]=1$. It follows that 
$sol(\Pi_{S_3}(\rho))$ contains a unique tuple whose  $X_i^{\ell-1}$-value is zero and whose $X_i^{\ell}$-value is one, namely the tuple $t'$. Therefore $t=t'$, which contradicts our assumption that $t\not\in rel(\rho)$. 
\item Assume that for some $0< \ell \leq r$, $t[X_i^{\ell-1}]=1$ while  $t[X_i^{\ell}]=0$. Observe that, by construction, $rel(\rho)$ does not contain a single tuple $t'$ for which 
$t'[X_i^{\ell-1}]=1$ while  $t'[X_i^{\ell}]=0$. In fact, $rel(\rho)$ was carefully constructed so that the bit values in the sequences $t'[X_i^0,X_i^1,\ldots,X_i^r]$  never decrease in any of its tuples. Therefore, in no relation $rel(c)$ of any constraint $c$  of $\Pi_{S_3}(\rho)$ where $scope(c)$ contains $X_i^{\ell-1},X_i^{\ell}$ and any other variable $X_j^u$, there is thus a tuple $f$ having $f[X_i^{\ell-1}]=1$ and $f[X_i^{\ell}]=0$. It follows that the join
$sol(\Pi_{S_3}(\rho))$ contains no tuple whose  $X_i^{\ell-1}$-value is one and whose $X_i^{\ell}$-value is zero. Contradiction.
\end{itemize}
We have thus established that for $1\leq i\leq n$, all bits of  $t[X_i^0,X_i^1,\ldots,X_i^r]$ must be  equal.  Let $\tau$ be the truth value assignment that for $1\leq i\leq n$ associates to each $p_i$ the truth value 
$t[X_i^0]=t[X_i^1]=\cdots=t[X_i^r]$. Let $C_h$ be any clause of $C$. Let the atoms of $C_h$ be $p_a, p_b$ and $p_c$. 
Define: 
\begin{quote}
$X_{(a)}:=X_a^0$ if $\tau(p_a)=1$ and $X_{(a)}:=X_a^r$ if 
 $\tau(p_a)=0$;\\
$X_{(b)}:=X_b^0$ if $\tau(p_b)=1$ and $X_{(b)}:=X_b^r$ if
 $\tau(p_b)=0$;\\
 $X_{(c)}:=X_c^0$ if $\tau(p_c)=1$ and $X_{(c)}:=X_c^r$, if 
 $\tau(p_c)=0$.
\end{quote}

Consider the constraint $q$ of $\Pi_{S_3}(\rho)$ having $\langle\XXa,\XXb,\XXc\rangle$ as scope. This constraint must have a tuple $t_q=\langle \tau(p_a),\tau(p_b),\tau(p_c) \rangle$, which is obviously 
identical to $t[\XXa,\XXb,\XXc]$. There is, therefore, a tuple $t'\in rel(\rho)$ such that 
$$t'[\XXa,\XXb,\XXc]= \langle\tau(p_a),\tau(p_b),\tau(p_c)\rangle.$$ 
 Given the specific values and positions of 
$\XXa$, $\XXb$, and $\XXc$ in $t'$, it is easily seen that the tuple $t'$ must belong to the group of {\em clause-induced tuples}, and more specifically,  $t'$ is induced by precisely clause $C_h$ and 
truth value assignment $\tau[p_a,p_b,p_c]$. To see this, let us
first recall that in our our encoding of a tuple identifier the first bit (i.e., bit 0) is always 0 and the last bit (i.e., bit $r$)  is always 1, which is {\em never} the case for the encoding of a truth value.  Now consider $\tau(p_a)$. If $\tau(p_a)=0$, then $t(\XXa)=t'(\XXa)=t'(X_a^r)=0$. If $\XXa=X_a^r$ were part of a tuple-identifier, then 
$t[\XXa]$, which is identical to $t[X_a^r]$,  could never 
have value zero, because, bit $r$ of a tuple identifier is always 1. Therefore, $\XXa$ must be part of a (representation of a) truth value assignment. Similarly, if $\tau(p_a)=1$, then $t(\XXa)=t'(\XXa)=t'(X_a^0)=1$. If $\XXa=X_a^0$ were part of a tuple-identifier, $t[\XXa]$ could never have value one, because all tuple identifiers have value zero at their bit position of index zero. 
Therefore, again,   $\XXa$ must be part of a (representation of a) truth value assignment. Exactly the same reasoning applies to $\XXb$ and $\XXc$. In summary, $t'$ is a tuple of $\rho$ that 
exactly describes truth value assignment $\tau$ restricted  to the three propositional variables $p_a,p_b$, and $p_c$. Given that these propositional variables jointly occur in clause $C_h$, $t'$ is a
clause-induced tuple, and $\tau$ is a ``legal" truth-value assignment that satisfies $C_h$. Given that $C_h$ was an arbitrary clause of $C$, $\tau$ satisfies all clauses of $C$, and thus  $C$ is satisfiable. 
We are done for $k=3$. The proof is easily modified to hold for any larger fixed value $k$. It suffices, for example, to start with $k$SAT instead of 3SAT. The proof goes through with 
the obvious adjustments to the numeric parameters. 
\end{proof}

\section{Proof of Theorem \ref{ident3} }~\label{AppendixC}

\vspace{-0.5cm}
\begin{xtheorem}{\ref{ident3}}
{\em For each fixed integer $k\geq 2$,  deciding 
for a single tri-valued constraint  $\rho$  
whether 
$sol(\Pi_{S_k}(\rho))=\rho$, that is,  whether $\rho$ is 
$k$-decomposable, is coNP-complete.} 
\end{xtheorem}

\begin{proof}  
For all constants $k$, the membership in coNP of our decision problem is already covered by (the upper bound in) Theorem~\ref{ident}. Moreover, the coNP-hardness for $k\geq 3$ is already proven in 
Theorem~\ref{ident2}, as bi-valued relations are trivially also $k$-valued relations (where the additional $k$-2  values appear in the domains but not in the actual constraint relations). Thus, what remains to be done is to prove coNP-hardness for $k=2$. 

\smallskip

We  use a transformation from 3COL from a graph $G=(V,E)$ as described in the proof of Theorem~\ref{ident} by applying similar vectorization techniques as in the proof of 
Theorem~\ref{ident2}. In particular, consider the relation $\rho$ obtained from the 3-colability network $\ncol$  in the hardness part of the proof of  Theorem~\ref{ident}, and let $s=|\rho|$ be the cardinality of $\rho$. Rather than transforming $\ncol$ (and thus the graph $G$) to $\rho$, we will transform it to a tri-valued constraint $\rho^*$  of the same cardinality $s$, that closely resembles $\rho$. To this aim,  for $1\leq i\leq n$, every scope variable $X_i$ of  $\ncol$ (and thus of $\rho$) is replaced by a block of $s+1$ variables $X_i^0,\ldots,X_i^s$,  which either encodes a color from $\{r,g,b\}$, or a tuple identifier. We  here use the following encoding: 
\begin{itemize}
\item Color {\em red} is encoded as a block consisting of $s+1$ consecutive positions having value $r$. 
\item Color {\em green} is encoded as a block consisting of  $s+1$  consecutive positions having value  $g$. 
\item Color {\em blue} is encoded by a leading $b$ (as an assignment to $X_i^0$ ) followed by a block containing $s$ consecutive positions having value $r$. 
\item The tuple identifier for tuple number $d$ is a block of length $s+1$ starting with a sequence of one or more $r$ elements, having a $b$ in the position corresponding to $X_i^d$, followed by $g$ elements. In other terms, this tuple identifier is a sequence of length $s+1$ of the form $r,\ldots,r,b,g,\ldots,g$, whose $d+1$st component is $b$.
\end{itemize}
The new relation $\rho^*$ thus has ${\mathit dom}(\rho^*)=\{r,g,b\}$ and 
$${\mathit scope}(\rho^*)=(X_1^0,X_1^1,\ldots,X_1^s,X_2^0,X_2^1,\ldots,X_2^s,\ldots\ldots, X_n^0,X_n^1,\ldots,X_n^s).$$ 
\medskip 
\noindent
CLAIM: $G$ is 3-colorable iff  $sol(\Pi_{S_2}(\rho^*)) - rel(\rho^*)$ is non-empty.

\smallskip

The {\em if-part} is not hard to see from our construction. In fact,  each correct graph coloring $\tau$ gives rise to a tuple $t$ in $sol(\Pi_{S_2}(\rho^*)) - rel(\rho^*)$ whose vectorized component  
$t[X_i^0,\ldots,X_i^s]$ representing vertex $v_i$ consists of the encoding of the color $\tau(v_i)$. 

\smallskip

Let us now prove the {\em only-if} part. Assume  there exists a tuple 
$t$ in $sol(\Pi_{S_2}(\rho^*)) - rel(\rho^*)$. We can show by similar arguments as  in the proof of Theorem~\ref{ident2} that $G$ must be 3-colorable. This is shown by the following successively derived facts:

\begin{enumerate}
\item Tuple $t$ can never have value  $b$ in an $X_i^j$-component with $j\neq 0$. In fact,  if it had a $b$ assigned to a variable $X_i^\ell$ with $\ell\neq 0$, this assignment would occur in a single tuple  $t'$ of $rel(\rho^*)$ only. Therefore in each relation $rel(c)$ of any constraint $c$ of 
$\Pi_{S_2}(\rho^*)$ where $scope(c)$ contains $X_i^\ell$ would contain a single tuple 
having $X_i^\ell=b$. But this means that the join of all relations $\Pi_{S_2}(\rho^*)$ contains a single tuple having $X_i^\ell=b$, namely $t'$ itself. But would imply $t=t'\in rel(\rho^*)$ which is a contradiction. 
\item No pair of consecutive values of any block $t[X_i^1,\ldots,X_i^s]$, for $1\leq i\leq n$ can coincide with $rg$ or $gr$. In fact, by construction, neither $rg$ or $gr$ occur as consecutive values in two consecutive 
columns labeled $X_i^\ell,X_i^{\ell+1}$, of $rel(\rho^*)$, where $\ell\geq 1$. Therefore, 
no relation $rel(c)$ of any constraint $c$ of $\Pi_{S_2}(\rho^*)$, whose scope is $X_i^\ell,X_i^{\ell+1}$,  where $\ell\geq 1$, contains tuple $rg$ or tuple $gr$. It follows that the 
join $sol(\Pi_{S_2}(\rho^*))$ cannot contain any tuple having  $rg$ or tuple $gr$ in consecutive 
components corresponding to  the variables (attributes) 
$X_i^\ell,X_i^{\ell+1}$, where $\ell\geq 1$. Given that $t\in sol(\Pi_{S_2}(\rho^*))$, the same follows for tuple $t$.
\item For each $1\leq i\leq n$, the block $t[X_i^1,\ldots,X_i^s]$ is made entirely of the same value, namely, either $r$ or $g$. This follows immediately  from the above facts 1 and 2. 

\item For $1\leq i\leq n$, each block of values $t[X_i^0,\ldots,X_i^s]$ precisely encodes  one of the colors {\em red}, {\em green}, or {\em blue}, according to our encoding scheme. 
To show this, it is sufficient to show that for $1\leq i\leq n$, if $t[X_i^1]=r$ then $t[X_i^0]\in\{r,b\}$, and if $t[X_i^1]=g$ then $t[X_i^0]=g$. This is shown just in the same way as Fact 2 above. By construction of $\rho^*$, 
the same property holds for each tuple of $\rho^*$, and thus for all the constraints with scope 
$\{ X_i^0,X_i^1 \}$ of $\Pi_{S_2}(\rho^*)$. Therefore,  the property must also hold for each tuple of the join $sol(\Pi_{S_2}(\rho^*))$ of $\Pi_{S_2}(\rho^*)$, and thus, 
in particular, for $t$. 

\item For each edge $\langle v_a,v_b\rangle \in E$, the blocks   $t[X_a^0,\ldots X_a^s]$  and    $t[X_b^0,\ldots X_b^s]$ represent {\em different} colors. To show this, define $\XXa:=X_a^s$ if $X_a^0=r$ and 
$\XXa:=X_a^0$ otherwise. Similarly, define $\XXb:=X_b^s$ if $X_b^0=r$ and 
$\XXb:=X_b^0$ otherwise. Let $q$ be the  constraint of $\Pi_{S_2}(\rho^*)$ with $scope(q)=\{\XXa,\XXb\}$. Clearly $t[\XXa,\XXb]=q[\XXa,\XXb]$. Thus there is a tuple $t'\in rel(\rho^*)$ such that 
$t[\XXa,\XXb]=t'[\XXa,\XXb]$. However, due to the particular value-position combinations, neither $t'[\XXa]$ nor $t'[\XXb]$ can be part of a tuple-identifier, and they  thus jointly represent a legal 
coloring of the edge $\langle v_a,v_b\rangle$ of $G$. Since  this is true for all edges 
$\langle v_a,v_b\rangle$ of $G$, all edges of $G$ are correctly colored by the coloring expressed 
by tuple $t$. 
\end{enumerate}
Therefore, $G$ is 3-colorable.  This concludes the proof of the {\em only-if} part of our claim, and thus the proof of our theorem.
\end{proof}

\medskip

\noindent{\bf Acknowledgments} Work funded by EPSRC Grant EP/G055114/1 ``Constraint
Satisfaction for Configuration: Logical Fundamentals, Algorithms, and Complexity". The author is a James Martin Senior Research Fellow. He thanks V.~B\'ar\'any, C.~Bessiere,  D.~Cohen, R.~Dechter,
 D.~Gaur, J.~Petke, M.~Vardi, M.~Yannakakis, S.~\v{Z}ivn\'{y}, and the referees of both the conference and the journal version for useful comments and/or pointers to earlier work.
\bibliographystyle{amsplain}
\vspace{-0.3cm}
\bibliography{AIJ-Gottlob}
\end{document}